\documentclass[letterpaper,11pt]{article}

\usepackage[margin=1in,letterpaper]{geometry} 
\usepackage[utf8]{inputenc} 
\usepackage[T1]{fontenc}    
\usepackage{hyperref}       
\usepackage{url}            
\usepackage{booktabs}       
\usepackage{amsfonts}       
\usepackage{nicefrac}       
\usepackage{microtype}      
\usepackage{xcolor}         

\usepackage[flushleft]{threeparttable}
\usepackage{bm}
\usepackage{graphicx}
\usepackage{subfigure}
\usepackage{algorithm}
\usepackage{algorithmic}
\usepackage{multirow}
\usepackage{amsmath}
\usepackage{amssymb}
\usepackage{amsthm}
\newtheorem{theorem}{Theorem}[section]

\newtheorem{proposition}[theorem]{Proposition}

\newtheorem{definition}[theorem]{Definition}

\title{Learning Graphon Autoencoders for \\Generative Graph Modeling}

\author{
  Hongteng Xu\textsuperscript{\rm 1, 2} \quad 
  Peilin Zhao\textsuperscript{\rm 3}\quad 
  Junzhou Huang\textsuperscript{\rm 3}\quad
  Dixin Luo\textsuperscript{\rm 4}\thanks{Corresponding author}\\
  \textsuperscript{\rm 1}\small{Gaoling School of Artificial Intelligence, Renmin University of China}\\
  \textsuperscript{\rm 2}\small{Beijing Key Laboratory of Big Data Management and Analysis Methods}\\
  \textsuperscript{\rm 3}\small{Tencent AI Lab}\\
  \textsuperscript{\rm 4}\small{School of Computer Science and Technology, Beijing Institue of Technology}\\
  \texttt{hongtengxu@ruc.edu.cn} \quad \texttt{dixin.luo@bit.edu.cn} \\
}

\begin{document}

\maketitle

\begin{abstract}
Graphon is a nonparametric model that generates graphs with arbitrary sizes and can be induced from graphs easily. 
Based on this model, we propose a novel algorithmic framework called \textit{graphon autoencoder} to build an interpretable and scalable graph generative model. 
This framework treats observed graphs as induced graphons in functional space and derives their latent representations by an encoder that aggregates Chebshev graphon filters. 
A linear graphon factorization model works as a decoder, leveraging the latent representations to reconstruct the induced graphons (and the corresponding observed graphs). 
We develop an efficient learning algorithm to learn the encoder and the decoder, minimizing the Wasserstein distance between the model and data distributions.
This algorithm takes the KL divergence of the graph distributions conditioned on different graphons as the underlying distance and leads to a reward-augmented maximum likelihood estimation.
The graphon autoencoder provides a new paradigm to represent and generate graphs, which has good generalizability and transferability.
\end{abstract}

\section{Introduction}
As a significant methodology for generative modeling, autoencoders map the data in the sample space $\mathcal{X}$ to a low-dimensional manifold embedded in a latent space $\mathcal{Z}\subset\mathbb{R}^C$.
Typically, an autoencoder is specified by an encoder $f:\mathcal{X}\mapsto\mathcal{Z}$ mapping the data to latent codes in $\mathcal{Z}$, a predefined or learnable prior distribution $p_\mathcal{Z}$ on $\mathcal{Z}$, and a decoder $h:\mathcal{Z}\mapsto\mathcal{X}$ mapping the latent codes back to $\mathcal{X}$. 
By learning these modules, the autoencoder minimizes the discrepancy between the data distribution $p_\mathcal{X}$ and the model distribution $p_{h(\mathcal{Z})}$~\cite{kingma2013auto,tolstikhin2018wasserstein}. 
Compared with other generative modeling strategies like generative adversarial networks (GANs)~\cite{goodfellow2014generative} and generative flows~\cite{kingma2018glow}, autoencoders can represent observed data explicitly in the latent space. 
Therefore, besides generating high-dimensional data like images~\cite{xu2020learning} and texts~\cite{wang2019topic}, autoencoders have been widely used to learn data representations for other downstream tasks, $e.g.$, data clustering and classification.

However, most existing autoencoders are designed for the data in the same space. 
They are often inapplicable for complicated structured data sampled from incomparable spaces, such as a collection of arbitrarily-sized unaligned graphs ($i.e.$, the graphs have different numbers of nodes and the correspondence between their nodes is unknown).
The variational graph autoencoder (VGAE)~\cite{kipf2016variational} and its variants~\cite{pan2018adversarially,wang2016structural} obtain node-level embeddings rather than a global graph representation. 
Recently, some models apply attention-based pooling layers to aggregate the node embeddings as the graph representation~\cite{vinyals2016order,li2016gated,luise2018differential}. 
However, these models often require side information to explore the clustering structure of the graphs, and they seldom consider the reconstructive and generative power of the graph representations.

\begin{figure}[t]
    \centering
    \includegraphics[width=0.95\linewidth]{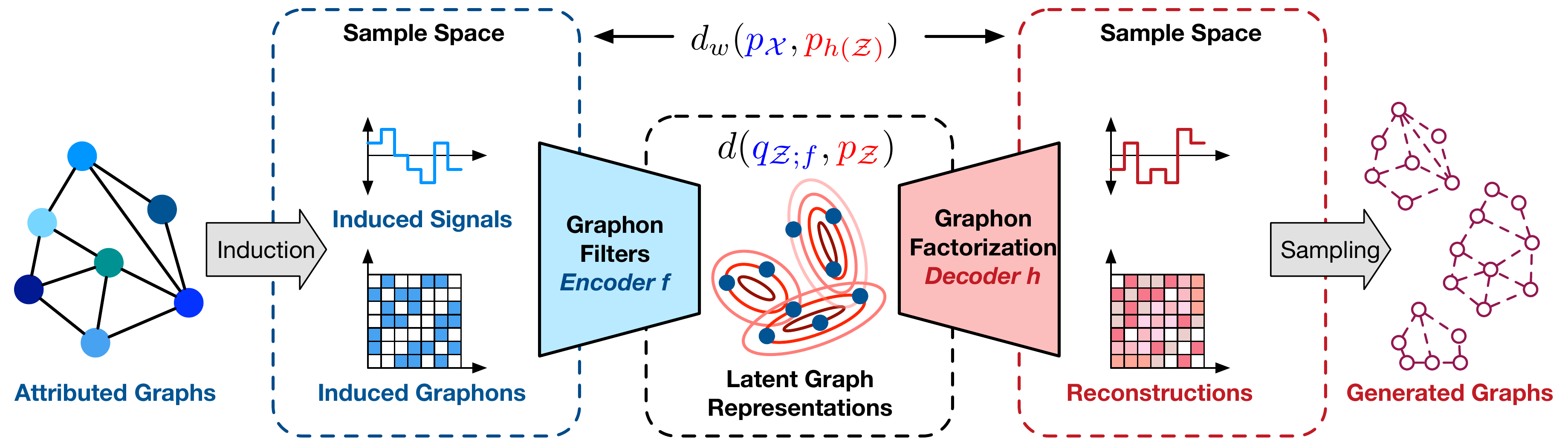}
    \caption{An illustration of our graphon autoencoder.}
    \label{fig:scheme}
\end{figure}

To overcome the challenges above, we propose a novel \textbf{Graphon Autoencoder (GNAE)}. 
Leveraging the theory of graphon~\cite{lovasz2012large}, we induce graphons ($i.e.$, two-dimensional symmetric Lebesgue measurable functions) from observed graphs and represent the node attributes associated with each graph as the signals defined on the graphons, such that the attributed graphs become the induced graphons and signals, which are in the same functional space.
As illustrated in Figure~\ref{fig:scheme}, our GNAE essentially achieves a Wasserstein autoencoder~\cite{tolstikhin2018wasserstein} for the graphons. 
The encoder of our GNAE is an aggregation of Chebyshev graphon filters, which can be implemented as a graph neural network (GNN) for the induced graphons. 
It outputs the latent representations of the induced graphons (or, equivalently, the observed graphs). 
The posterior distribution of the latent representations is regularized by their prior distribution. 
The decoder of our GNAE is a graphon factorization model, which can reconstruct graphons from the latent representations and sample graphs with arbitrary sizes. 
For each induced graphon, its latent representation corresponds to the coefficients of the graphon factors in the decoder, which explicitly indicates its similarity to the factors. 
Therefore, our GNAE achieves an interpretable and scalable generative model for graphs. 

We develop an efficient algorithm to achieve our GNAE.
For each reconstructed graphon, we sample graphs and calculate their distributions conditioned on the reconstructed graphon and the input graphon, respectively. 
Taking the KL divergence between the conditional distributions as the underlying distance, we minimize the Wasserstein distance between the data and model distributions and learn our GNAE by a reward-augmented maximum likelihood (RAML) estimation method~\cite{norouzi2016reward}. 
This algorithm avoids dense matrix multiplications and the backpropagation corresponding to the fused Gromov-Wasserstein (FGW) distance~\cite{vayer2020fused} between graphons, which owns low computational complexity. 
Experiments show that our GNAE performs well on representing and generating graphs, which have good generalizability ($i.e.$, generating graphs with various sizes but similar structures) and transferability ($i.e.$, training on a graph set and testing on others).

\section{Proposed Model}
\subsection{From attributed graphs to graphons with signals}
Mathematically, a graphon is a two-dimensional symmetric Lebesgue measurable function, denoted as $g:\Omega^2\mapsto [0, 1]$, where $\Omega$ is a measure space with a probability measure $\mu_{\Omega}$. 
Typically, we often set $\Omega=[0, 1]$ and $\mu_{\Omega}$ as a uniform distribution on $\Omega$. 
Associated with the graphon, we can define a $M$-dimensional signal on it~\cite{morency2021graphon}, which is denoted as $s:\Omega\mapsto \mathbb{R}^M$. 

Denote the graphon space as $\mathcal{G}$ and the signal space as $\mathcal{S}$, respectively. 
For arbitrary $g_1,g_2\in\mathcal{G}$, their $\delta_p$ distance~\cite{lovasz2012large} is $\delta_p(g_1, g_2):=\inf_{\phi \in \mathcal{F}_{\Omega}}\|g_1 - g_2^{\phi}\|_{p}$, where $\|g\|_{p}:=(\int_{\Omega^2} | g(u, v)|^p dudv)^{\frac{1}{p}}$. 
Typically, we set $p=1$ or $2$. 
If $\delta_{p}(g_1, g_2)=0$, we say $g_1$ and $g_2$ are equivalent, denoted as $g_1\cong g_2$. 
The work in~\cite{borgs2008convergent,gao2019graphon} shows that the quotient space $(\widehat{\mathcal{G}}, \delta_{p})$, where $\widehat{\mathcal{G}}:=\mathcal{G}\setminus \cong$, is homomorphic. 
$\delta_{p}$ is widely used in practice because of its computability. 
Especially, the work in~\cite{janson2013graphons,xu2021learning} indicates that the $\delta_p$ distance is equivalent to the order-$p$ Gromov-Wasserstein (GW) distance~\cite{memoli2011gromov}:
\begin{definition}
For arbitrary $g_1,g_2\in\mathcal{G}$, their order-$p$ Gromov-Wasserstein distance is
\begin{eqnarray}
\begin{aligned}
d_{\text{gw}}(g_1, g_2):=\sideset{}{_{\pi\in\Pi(\mu_{\Omega},\mu_{\Omega})}}\inf\Bigl(\int_{\Omega^2\times\Omega^2}|g_1(u,u')-g_2(v,v')|^p d\pi(u,v)d\pi(u',v')\Bigr)^{\frac{1}{p}},
\end{aligned}
\end{eqnarray}
where $\Pi(\mu_{\Omega},\mu_{\Omega}):=\{\pi\geq 0 | \int_{u\in\Omega}d\pi(u,v)=\mu_{\Omega}, \int_{v\in\Omega}d\pi(u,v)=\mu_{\Omega}\}$. 
\end{definition}

For $\mathcal{S}$, we can apply the Wasserstein distance as its metric:
\begin{definition}
For arbitrary $s_1,s_2\in\mathcal{S}$, their order-$p$ Wasserstein distance is
\begin{eqnarray}\label{eq:d_w}
\begin{aligned}
d_{\text{w}}(s_1, s_2) := \sideset{}{_{\pi\in\Pi(\mu_{\Omega},\mu_{\Omega})}}\inf\Bigl(\int_{\Omega^2}\|s_1(u)-s_2(v)\|_p^p d\pi(u,v)\Bigr)^{\frac{1}{p}}.
\end{aligned}
\end{eqnarray}
\end{definition}

\textbf{Sampling graphs:} 
Graphon is a nonparametric graph generative model. 
We can sample graphs with arbitrary sizes from a graphon by the following steps:
\begin{eqnarray}\label{eq:generate_graph}
\begin{aligned}
\text{1)}~\text{for}~n=1,..,N,~v_n\sim \mu_{\Omega};~\text{2)}~a_{nn'}\sim \text{Bernoulli}(g(v_n, v_{n'}));~\text{3)}~\bm{s}_n\sim \mathcal{N}(s(v_n), \sigma).
\end{aligned}
\end{eqnarray}
The first step is sampling $N$ nodes independently from $\mu_{\Omega}$. 
The second step generates an adjacency matrix $\bm{A}=[a_{nn'}]\in \{0, 1\}^{N\times N}$, whose elements are sampled from the Bernoulli distributions determined by the graphon. 
When a signal is available, we can sample the attributes associated with the nodes, denoted as $\bm{S}=[\bm{s}_n]\in\mathbb{R}^{N\times M}$, from the distributions determined by the signal, $e.g.$, the Gaussian distributions in (\ref{eq:generate_graph}).
For convenience, we denote $G(\bm{A}, \bm{S})$ as the sampled graph. 

\textbf{Inducing graphons:} 
We induce a graphon and a signal from an attributed graph as follows.
\begin{definition}[Induced Graphon]\label{def:step}
For a graph $G(\bm{A},\bm{S})$, where $\bm{A}=[a_{nn'}]\in \{0, 1\}^{N\times N}$ and $\bm{S}=[\bm{s}_n]\in\mathbb{R}^{N\times M}$, we can induce a graphon and its corresponding signal as two step functions:
\begin{eqnarray}\label{eq:induced_graphon}
\begin{aligned}
g_{\mathcal{P}}(v,v')=\sideset{}{_{n,n'=1}^{N}}\sum a_{nn'}1_{\mathcal{P}_n}(v)1_{\mathcal{P}_{n'}}(v'),~\text{and}~s_{\mathcal{P}}(v)=\sideset{}{_{n=1}^{N}}\sum \bm{s}_{n}1_{\mathcal{P}_n}(v),~
\forall~v,v'\in\Omega,
\end{aligned}
\end{eqnarray}
where $\mathcal{P}=\{\mathcal{P}_n\}_{n=1}^{N}$ represents $N$ equitable partitions of $\Omega$, $i.e.$, $\cup_{n}\mathcal{P}_n=\Omega$ and $|\mathcal{P}_n|=|\mathcal{P}_{n'}|$ for all $n\neq n'$. 
The indicator $1_{\mathcal{P}_n}(v)=1$ if $v\in\mathcal{P}_{n}$, otherwise it equals to $0$. 
\end{definition}
Obviously, $g_{\mathcal{P}}\in\mathcal{G}$ and $s_{\mathcal{P}}\in\mathcal{S}$.  
The step function approximation lemma~\cite{chan2014consistent} shows that for the graphs sampled from a graphon $g$, the average of their induced graphons provides a consistent estimation of $g$. 
The estimation error reduces with the increase of the number and the size of the graphs.

\subsection{A graphon autoencoder in functional space}
The graphons and their associated signals, denoted as $\{(g,s)\}$, can be viewed as samples in a functional space $(\mathcal{X},d_{\mathcal{X}},\mathbb{P})$. 
Here, $\mathcal{X}=\mathcal{G}\times \mathcal{S}$, $d_{\mathcal{X}}$ is an underlying distance defining the discrepancy between different samples,\footnote{Note that the underlying distance may not be a strict metric in practice.} which will be introduced below and discussed in-depth in Section~\ref{sec:learning}, and $\mathbb{P}$ represents the set of probability measures defined on $\mathcal{X}$. 

By inducing graphons with signals, we can represent the arbitrarily-sized unaligned graphs as the samples in the same space. Accordingly, existing machine learning techniques like autoencoders become applicable. 
In particular, our graphon autoencoder (GNAE) can be viewed as a Wasserstein autoencoder of the graphons, which consists of an encoder $f:\mathcal{X}\mapsto\mathcal{Z}$, a decoder $h:\mathcal{Z}\mapsto\mathcal{X}$, and a learnable latent prior distribution $p_{\mathcal{Z}}$. 
Given attributed graphs, we obtain a set of induced graphons and associated signals, denoted as $\{(g_{\mathcal{P}}, s_{\mathcal{P}})\}$ and learn the autoencoder to minimize the order-1 Wasserstein distance between the (unknown) data distribution $p_{\mathcal{X}}$ and the model distribution $p_{h(\mathcal{Z})}$, $i.e.$, $\min d_{\text{w}}(p_\mathcal{X}, p_{h(\mathcal{Z})})$, where $p_\mathcal{X},p_{h(\mathcal{Z})}\in\mathbb{P}$.
According to Theorem 1 in~\cite{tolstikhin2018wasserstein}, we relax the optimization problem as
\begin{eqnarray}\label{eq:wae}
\begin{aligned}
\sideset{}{_{f,h,p_{\mathcal{Z}}}}\min~\mathbb{E}_{\bm{x}\sim p_\mathcal{X}}\mathbb{E}_{\bm{z}\sim q_{\mathcal{Z}|\mathcal{X}; f}}[d_{\mathcal{X}}(\bm{x}, h(\bm{z}))] + \gamma d(q_{\mathcal{Z};f},p_\mathcal{Z}),
\end{aligned}
\end{eqnarray}
where each $\bm{x}=(g_{\mathcal{P}},s_{\mathcal{P}})$ represents a tuple of the graphon and signal induced from the corresponding observed graph. 
$d_{\mathcal{X}}(\bm{x},h(\bm{z}))$ represents the reconstruction error of the sample $\bm{x}$.
$q_{\mathcal{Z};f}=\mathbb{E}_{\bm{x}\sim p_\mathcal{X}}[q_{\mathcal{Z}|\bm{x}; f}]$ is the expected latent posterior conditioned on different samples, which is required to be closed to the latent prior $p_{\mathcal{Z}}$ under the metric $d$.
Parameter $\gamma$ achieves a trade-off between reconstruction loss and the regularizer. 
As shown in~(\ref{eq:wae}), our GNAE learns the encoder, the decoder, and the latent prior distribution jointly. 
These modules are implemented as follows.

\begin{figure}[t]
    \centering
    \subfigure[Aggregated graphon filters (encoder)]{
    \includegraphics[height=3.4cm]{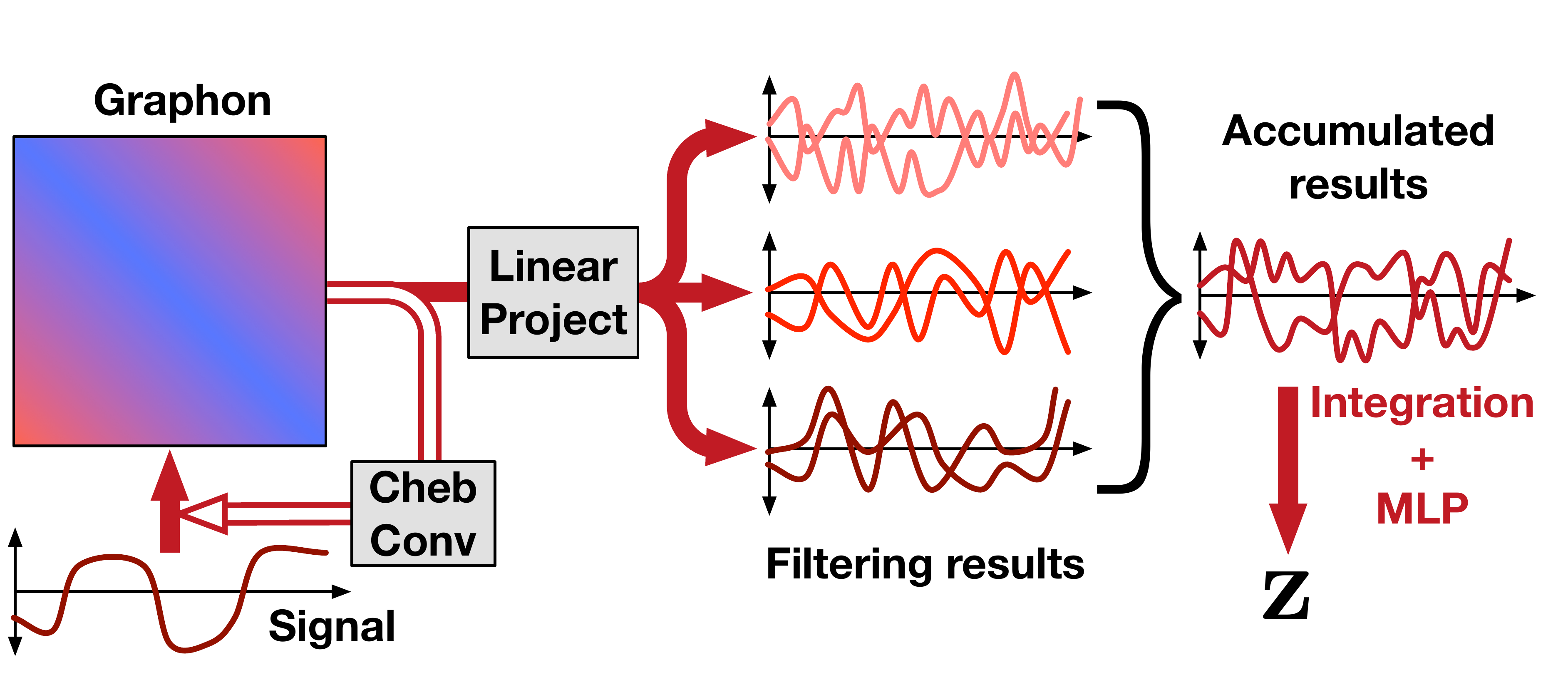}\label{fig:encoder}
    }
    \subfigure[Graphon factorization model (decoder)]{
    \includegraphics[height=3.4cm]{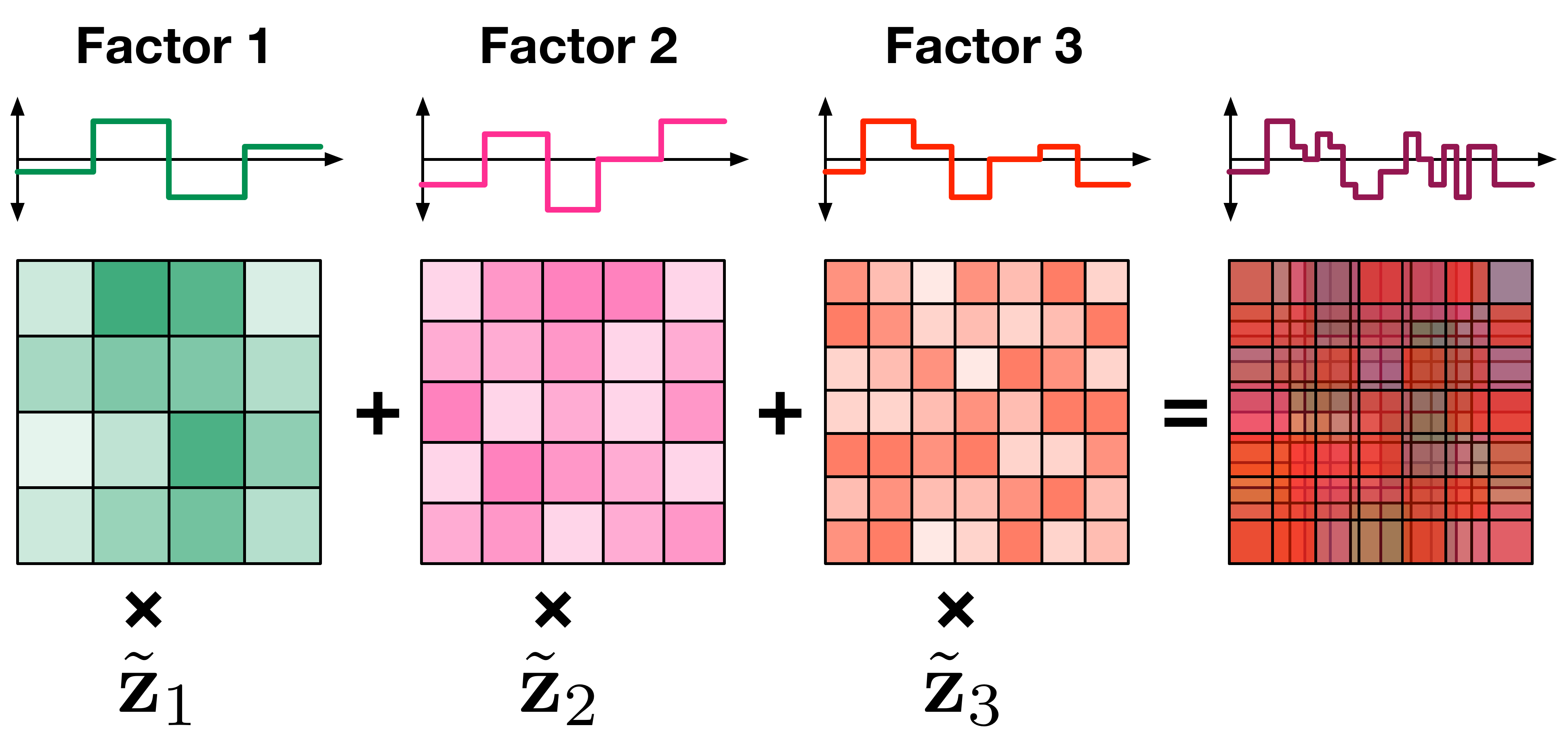}\label{fig:decoder}
    }
    \caption{Illustrations of the encoder and the decoder of our GNAE.}
\end{figure}

\textbf{Latent prior distribution} Following the work in~\cite{xu2020learning}, we set the latent prior $p_{\mathcal{Z}}$ as a learnable Gaussian mixture model (GMM) and implement the regularizer in (\ref{eq:wae}) as the sliced fused Gromov-Wasserstein (SFGW) distance, $i.e.$, $d(q_{\mathcal{Z};f},p_\mathcal{Z}):=d_{\text{sfgw}}(q_{\mathcal{Z};f},p_\mathcal{Z})$. 
This configuration helps us to learn latent representations with clustering structures. 

\textbf{Encoder} The encoder of our GNAE is designed as an aggregation of Chebyshev graphon filters. 
Given a graphon $g$ and its corresponding signal $s$, we achieve our encoder as follows:
\begin{eqnarray}\label{eq:filter}
\begin{aligned}
&\bm{z} = f((g,s))=\text{MLP}\Bigl(\int_{\Omega}\sideset{}{_{j=0}^{J}}\sum \theta_j(s^{(j)}(v))dv\Bigr),~\text{where}\\
&s^{(0)}(v) = s(v),~s^{(1)}(v) = L_g(s^{(0)}(v)) =  \int_{\Omega}g(u,v) (s^{(0)}(v) - s^{(0)}(u))du,~\text{and}\\
&s^{(j)}(v) = 2L_g(s^{(j-1)}(v))-s^{(j-2)}(v)~\text{for}~j>1.
\end{aligned}
\end{eqnarray}
In the $j$-th step, the filtering result $s^{(j)}$ is obtained by applying a Laplacian filter to $s^{(j-1)}$ ($i.e.$, $L_g(s^{(j-1)}(v))$) and treating $s^{(j-2)}$ as an offset.
$\theta^{(j)}(\cdot)$ is a linear projection, mapping each $s^{(j)}(v)$ to $\mathbb{R}^{D}$. 
We obtain the aggregated filtering result by accumulating and integrating the results of different steps. 
Finally, we apply a multi-layer perceptron (MLP) network to derive a $C$-dimensional latent representation. 
Figure~\ref{fig:encoder} illustrates the scheme of our encoder.

For general graphons and signals, we can implement the filtering process above based on Fourier transform~\cite{bracewell1966fourier}, whose complexity is high.
Fortunately, for the induced graphon and signal $(g_{\mathcal{P}},s_{\mathcal{P}})$, we can implement the graphon filters by a Chebyshev spectral graph convolutional (ChebConv) network~\cite{defferrard2016convolutional}. 
For $j>0$, if $\Omega=[0,1]$ and $\mathcal{P}=\{\mathcal{P}_n\}_{n=1}^{N}$ are equitable partitions, we have
\begin{eqnarray}\label{eq:gcn}
\begin{aligned}
\int_{\Omega}g_{\mathcal{P}}(u,v) (s_{\mathcal{P}}^{(j-1)}(v) - s_{\mathcal{P}}^{(j-1)}(u))du
=\frac{1}{N}\sideset{}{_{n=1}^{N}}\sum(\bm{L}\bm{S}^{(j-1)})_{n}1_{\mathcal{P}_n}(v),~\text{for}~v\in\Omega,
\end{aligned}
\end{eqnarray}
where $\bm{L}=\text{diag}(\bm{A1})-\bm{A}$ is the Laplacian graph matrix, $\bm{S}^{(j-1)}=[\bm{s}_n^{(j-1)}]$ is a matrix of the $(j-1)$-th signal, and each $\bm{s}_n^{(j-1)}$ corresponds to the signal in the partition $\mathcal{P}_n$. 
Accordingly, $(\bm{L}\bm{S}^{(j-1)})_{n}$ is the $n$-th row of $\bm{L}\bm{S}^{(j-1)}$.
Plugging (\ref{eq:gcn}) into (\ref{eq:filter}), we can derive the latent representation as $\bm{z}=\text{MLP}(\sum_{n=1}^{N}(\sum_{j=0}^{J}\frac{1}{N^{j+1}}\theta_j(\bm{S}^{(j)}))$, which can be implemented as a ChebConv network followed by an average pooling layer and a MLP.

\textbf{Decoder}
For each $\bm{z}=[z_1,..,z_C]$ derived from the encoder, we apply a factorization model to reconstruct the corresponding graphon and signal. 
Specifically, the decoder consists of $C$ graphon factors, denoted as $\{(\tilde{g}_c, \tilde{s}_c)\}_{c=1}^{C}$.
Each graphon factor corresponds to two step functions defined as (\ref{eq:induced_graphon}) shows. 
Accordingly, the reconstructed sample can be represented as $h(\bm{z})=(\hat{g},\hat{s})$, where
\begin{eqnarray}\label{eq:reconstruct}
\begin{aligned}
\hat{g}=\sideset{}{_{c=1}^{C}}\sum \tilde{z}_c \tilde{g}_c,~\text{and}~\hat{s}=\alpha(\sideset{}{_{c=1}^{C}}\sum \tilde{z}_c \tilde{s}_c),~\text{with}~\tilde{z}_c =\text{softmax}_c(\bm{z})=\frac{\exp(z_c)}{\sum_{c'}\exp(z_{c'})}.
\end{aligned}
\end{eqnarray}
We reparameterize each $\tilde{g}_c$ as $\sigma(b_c)$, where $b_c(u,v):\Omega^2\mapsto \mathbb{R}$ is a step function with unbounded output range and $\sigma(\cdot)$ is a sigmoid function, and let the latent representation pass through a softmax layer. 
This setting makes $\{\tilde{g}_c\}_{c=1}^{C}$ and $\hat{g}$ in the space $\mathcal{G}$. 
The function $\alpha(\cdot)$ depends on the type of the original signal, which can be ReLU, softmax, sigmoid, $etc$. 
As shown in Figure~\ref{fig:decoder}, the graphon factors may have different partitions.
Applying the inclusion-exclusion principle~\cite{jukna2011extremal}, we have
\begin{proposition}\label{prop:partition}
Suppose that $\{\tilde{g}_c:[0,1]^2\mapsto [0,1]\}_{c=1}^{C}$ are 2D step functions, each of which has $N_c$ equitable partitions. 
Denote $\{\mathcal{L}_c\}_{c=1}^{C}$ as the sets of the landmarks indicating the partitions, where $\mathcal{L}_c=\{\frac{1}{N_c},...,\frac{N_c-1}{N_c}\}$. 
For the $\hat{g}$ derived by (\ref{eq:reconstruct}), the number of its partitions is
$|\mathcal{P}|=|\cup_{c=1}^{C}\mathcal{L}_c| + 1=\sum_{\emptyset\neq \mathcal{C}\subset\{1,..,C\}} (-1)^{|\mathcal{C}|}|\cap_{c\in\mathcal{C}}\mathcal{L}_c| + 1$.
If all the $N_c$'s are prime numbers, $|\mathcal{P}|=\sum_{c=1}^{C}|\mathcal{L}_c|+1$.
\end{proposition}
Proposition~\ref{prop:partition} shows the number of the partitions of the reconstructed graphon can be much larger than that of the graphon factors, which is beneficial for the capability of our model.

\section{Learning algorithm}\label{sec:learning}
\subsection{The fused Gromov-Wasserstein distance between graphons}\label{ssec:distance}
Besides the modules above, the key of our GNAE is the underlying distance $d_{\mathcal{X}}$. 
A straightforward way is implementing $d_{\mathcal{X}}$ as the fused Gromov-Wasserstein (FGW) distance~\cite{vayer2020fused}:
\begin{definition}
For $\bm{x}_1,\bm{x}_2\in\mathcal{X}$, where $\bm{x}_1=(g_1,s_1)$ and $\bm{x}_2=(g_2,s_2)$, their order-$p$ fused Gromov-Wasserstein distance, denoted as $d_{\text{fgw}}(\bm{x}_1, \bm{x}_2)$, is 
\begin{eqnarray*}
\begin{aligned}
\inf_{\pi\in\Pi(\mu_{\Omega},\mu_{\Omega})}\Bigl(\int_{\Omega^2\times\Omega^2}|g_1(u,u')-g_2(v,v')|^p d\pi(u,v)d\pi(u',v')+\int_{\Omega^2}\|s_1(u)-s_2(v)\|_p^p d\pi(u,v)\Bigr)^{\frac{1}{p}},
\end{aligned}
\end{eqnarray*}
\end{definition}
The FGW distance combines the GW distance between graphons and the Wasserstein distance between signals, enforcing them share the same optimal transport $\pi(u,v)$. 
It is a metric for the quotient space $\widehat{\mathcal{X}}:=\mathcal{X}\setminus \cong$ when $p=1$ and a semi-metric when $p>1$~\cite{vayer2020fused}. 

We can compute the FGW distance by solving an optimization problem with finite variables when dealing with the graphons and signals formulated as step functions. 
\begin{proposition}\label{prop1}
Given $\bm{x}_{1,\mathcal{P}}=(g_{1,\mathcal{P}},s_{1,\mathcal{P}})$ and $\bm{x}_{2,\mathcal{Q}}=(g_{2,\mathcal{Q}},s_{2,\mathcal{Q}})$, where
\begin{eqnarray*}
&g_{1,\mathcal{P}}(v,v')=\sideset{}{_{n,n'=1}^{N}}\sum g_{1,nn'}1_{\mathcal{P}_n}(v)1_{\mathcal{P}_{n'}}(v'),~g_{2,\mathcal{Q}}(v,v')=\sideset{}{_{m,m'=1}^{M}}\sum g_{2,mm'}1_{\mathcal{Q}_m}(v)1_{\mathcal{Q}_{m'}}(v'),\\
&s_{1,\mathcal{P}}(v)=\sideset{}{_{n=1}^{N}}\sum\bm{s}_{1,n} 1_{\mathcal{P}}(v),~s_{2,\mathcal{Q}}(v)=\sideset{}{_{m=1}^{M}}\sum\bm{s}_{2,m} 1_{\mathcal{Q}}(v)
\end{eqnarray*}
are step functions, we have
\begin{eqnarray}\label{eq:d_fgw2}
\begin{aligned}
d_{\text{fgw}}(\bm{x}_{1,\mathcal{P}}, \bm{x}_{2,\mathcal{Q}})=\sideset{}{_{\bm{T}\in\Pi(\bm{\mu}_{\mathcal{P}},\bm{\mu}_{\mathcal{Q}})}}\min ( \langle\bm{D}_g, \bm{T}\otimes\bm{T}\rangle + \langle\bm{D}_s, \bm{T}\rangle )^{\frac{1}{p}},
\end{aligned}
\end{eqnarray}
where $\bm{D}_g=[|g_{1,nn'}-g_{2,mm'}|^p]\in\mathbb{R}^{N^2\times M^2}$, $\bm{D}_s=[\|\bm{s}_{1,n}-\bm{s}_{2,m}\|_p^p]\in\mathbb{R}^{N\times M}$, $\otimes$ represents Kronecker product, and $\Pi(\bm{\mu}_{\mathcal{P}},\bm{\mu}_{\mathcal{Q}})=\{\bm{T}\geq\bm{0}|\bm{T}\bm{1}=\bm{\mu}_{\mathcal{P}},\bm{T}^{\top}\bm{1}=\bm{\mu}_{\mathcal{Q}}\}$ with $\bm{\mu}_{\mathcal{P}}=[\frac{|\mathcal{P}_1|}{|\Omega|},..,\frac{|\mathcal{P}_N|}{|\Omega|}]$ and $\bm{\mu}_{\mathcal{Q}}=[\frac{|\mathcal{Q}_1|}{|\Omega|},..,\frac{|\mathcal{Q}_M|}{|\Omega|}]$.
\end{proposition}

Although (\ref{eq:d_fgw2}) is computable, its computational complexity is so high as $\mathcal{O}(N^4)$  for the graphons with $N$ partitions (or $\mathcal{O}(N^3)$ if $p=2$~\cite{peyre2016gromov}). 
What is worse, because the graphons reconstructed by our GNAE are non-sparse 2D step functions, some commonly-used acceleration strategies like the sliced FGW distance~\cite{xu2020learning} and the sparse matrix multiplications~\cite{titouan2019optimal,xu2019scalable} become inapplicable. 
Additionally, the underlying distance is parameterized by the GNAE model, so we have to consider the gradient of the optimal transport matrix $\bm{T}$ to the model parameters, which is expensive on both time and memory~\cite{xie2020hypergradient}.

\subsection{The KL divergence between conditional graph distributions}\label{ssec:raml}
Facing the issues above, we need to explore a surrogate of the FGW distance when learning our GNAE model.
A competitive choice is the KL divergence between the distributions of graphs conditioned on $h(\bm{z})$ and $\bm{x}$, denoted as $d_{\text{KL}}(q(G|\bm{x}), p(G|h(\bm{z})))$. 
Specifically, given a reconstructed sample, $i.e.$, $h(\bm{z})=(\hat{g},\hat{s})$, we sample a set of attributed graphs, denoted as $\mathcal{Y}=\{G(\widehat{\bm{A}},\widehat{\bm{S}})\}$. 
Each $G(\widehat{\bm{A}},\widehat{\bm{S}})$ has $K$ nodes.
According to (\ref{eq:generate_graph}), the likelihood of $G(\widehat{\bm{A}},\widehat{\bm{S}})$ is 
\begin{eqnarray}\label{eq:pG}
\begin{aligned}
&p(G|h(\bm{z})) =p(\mathcal{V})p(\widehat{\bm{A}}|\hat{g},\mathcal{V})p(\widehat{\bm{S}}|\hat{s},\mathcal{V})
=\frac{1}{|\Omega|^{K}}\sideset{}{_{k,k'=1}^{K}}\prod p(\hat{a}_{kk'}|\hat{g}(v_k,v_{k'}))
\sideset{}{_{k=1}^{K}}\prod p(\hat{\bm{s}}_{k}|\hat{s}(v))
\\
&\propto\sideset{}{_{k,k'=1}^{K}}\prod \hat{g}(v_{k},v_{k'})^{\hat{a}_{kk'}}(1-\hat{g}(v_{k},v_{k'}))^{1-\hat{a}_{kk'}}\sideset{}{_{k=1}^{K}}\prod \exp\Bigl(-\frac{\|\hat{\bm{s}}_{k} - \hat{s}(v_k)\|_2^2}{2M\sigma^2}\Bigr),
\end{aligned}
\end{eqnarray}
where $\mathcal{V}=\{v_1,..,v_K\}$ and each $v_k$ is sampled independently from $\mu_{\Omega}$, so that $p(\mathcal{V})=\frac{1}{|\Omega|^{K}}$ when $\mu_{\Omega}$ is a uniform distribution.

Additionally, we leverage an exponentiated payoff distribution~\cite{norouzi2016reward} to approximate the probability of each $G(\widehat{\bm{A}},\widehat{\bm{S}})$ conditioned on the observed graphon $\bm{x}$:
\begin{eqnarray}\label{eq:qG}
\begin{aligned}
q(G|\bm{x}) = \frac{\exp(r(\hat{\bm{x}}_{G},\bm{x}))}{\sum_{G'\in\mathcal{Y}}\exp(r(\hat{\bm{x}}_{G'},\bm{x}))}=\frac{\exp(-d_{\text{fgw}}(\hat{\bm{x}}_{G},\bm{x}) / \tau)}{\sum_{G'\in\mathcal{Y}}\exp(-d_{\text{fgw}}(\hat{\bm{x}}_{G'},\bm{x}) / \tau)},
\end{aligned}
\end{eqnarray}
where $\hat{\bm{x}}_{G}$ is the graphon (and the signal) induced from the graph $G$, and $r(\hat{\bm{x}}_{G},\bm{x})=-\frac{d_{\text{fgw}}(\hat{\bm{x}}_{G},\bm{x})}{\tau}$ is the reward function implemented as the negative order-2 FGW distance between $\hat{\bm{x}}_{G}$ and $\bm{x}$. 
Parameter $\tau$ controls the smoothness of $q(G|\bm{x})$. 
In our work, we set $\tau$ adaptively as $\min_{G\in\mathcal{Y}} d_{\text{fgw}}(\hat{\bm{x}}_G,\bm{x})$. 

\subsection{Reward-augmented maximum likelihood estimation}
Leveraging $d_{\text{KL}}(q(G|\bm{x}), p(G|h(\bm{z})))$, we develop an efficient learning algorithm with much lower computational complexity. 
Specifically, we can rewrite $d_{\text{KL}}(q(G|\bm{x}), p(G|h(\bm{z})))$ as
\begin{eqnarray}\label{eq:kld}
\begin{aligned}
d_{\text{KL}}(q(G|\bm{x}), p(G|h(\bm{z})))=-\mathbb{E}_{G\sim q(G|\bm{x})}[\log p(G|h(\bm{z}))] + \mathbb{E}_{G\sim q(G|\bm{x})}[\log q(G|\bm{x})],
\end{aligned}
\end{eqnarray}
where the second term is the entropy of the sampled graphs, which is a constant with respect to the model.
Plugging (\ref{eq:kld}) into (\ref{eq:wae}), we learn our GNAE by
\begin{eqnarray}\label{eq:gnae}
\begin{aligned}
\sideset{}{_{f,h,p_{\mathcal{Z}}}}\min-\mathbb{E}_{\bm{x}\sim p_\mathcal{X}}\mathbb{E}_{\bm{z}\sim q_{\mathcal{Z}|\mathcal{X}; f}}\mathbb{E}_{G\sim q(G|\bm{x})}[\log p(G|h(\bm{z}))] + \gamma d_{\text{sfgw}}(q_{\mathcal{Z};f},p_\mathcal{Z}).
\end{aligned}
\end{eqnarray}
This optimization problem can be solved by the reward-augmented maximum likelihood (RAML) method~\cite{norouzi2016reward}. 
The scheme of our learning algorithm is shown in Algorithm~\ref{alg:raml}. 
In principle, when the sampled graph is close to the original input graph, the FGW distance between their induced graphon will be small. 
Accordingly, the log-likelihood of the graph ($i.e.$, $\log p(G|h(\bm{z}))$) will be assigned to a large weight ($i.e.$, $q(G|\bm{x})$), and the model will be updated to increase the likelihood. 

\begin{algorithm}[t]
    \caption{Learning a GNAE by RAML}	
    \label{alg:raml}
	\begin{algorithmic}[1]
	    \INPUT A set of graphons and signals induced from observed attributed graphs, denoted as $\mathcal{X}$. 
	    \OUTPUT An encoder $f$, a decoder $h$, and a latent prior  $p_{\mathcal{Z}}(\bm{z})=\frac{1}{T}\sum_t\mathcal{N}(\bm{z};\bm{\mu}_{t},\text{diag}(\bm{\sigma}_t^2))$.
	    \STATE \textbf{for} each epoch
	    \STATE \quad\textbf{for} each batch $\{\bm{x}_n\}_{n=1}^{N_b}\subset \mathcal{X}$
	    \STATE \quad\quad\textbf{for} $n=1,..,N_b$
	    \STATE \quad\quad\quad Samples of $q_{\mathcal{Z};f}$: $\bm{z}_n=f(\bm{x}_n)$.
	    \STATE \quad\quad\quad Samples of $p_{\mathcal{Z}}$: $t\sim \text{Categorical}(\frac{1}{T})$, and $\bm{z}'_n\sim \mathcal{N}(\bm{\mu}_{t},\text{diag}(\bm{\sigma}_t^2))$.
	    \STATE \quad\quad\quad Sample attributed graphs $\{G_i(\widehat{\bm{A}}_i,\widehat{\bm{S}}_i)\}_{i=1}^{I}$ from $h(\bm{z}_n)$ and induce $\{\hat{\bm{x}}_{G_i}\}_{i=1}^{I}$ by (\ref{eq:induced_graphon}).
	    \STATE \quad\quad\quad Compute $\{d_{\text{fgw}}(\hat{\bm{x}}_{G_i},\bm{x}_n)\}_{i=1}^{I}$ by (\ref{eq:d_fgw2}) and obtain $\{q(G_i|\bm{x}_n)\}_{i=1}^{I}$ by (\ref{eq:qG}).
	    \STATE \quad\quad\quad Compute $\{p(G_i|h(\bm{z}_n))\}_{i=1}^{I}$ by (\ref{eq:pG}).
	    \STATE \quad\quad Calculate $\mathcal{L}=-\sum_{n=1}^{N_b}\sum_{i=1}^{I}q(G_i|\bm{x}_n)\log p(G_i|h(\bm{z}_n)) + \gamma d_{\text{sfgw}}(q_{\mathcal{Z};f}, p_{\mathcal{Z}})$.
	    \STATE \quad\quad Update the model by Adam optimizer~\cite{kingma2014adam}.
	\end{algorithmic}
\end{algorithm}

Although our learning algorithm still involves computing FGW distances, it has obvious advantages on computational complexity compared to using $d_{\text{fgw}}(\bm{x},h(\bm{z}))$ as underlying distance directly. 
Firstly, the number of each sampled graph's nodes ($i.e.$, $K$) can be much smaller than that of the reconstructed graphon's partitions ($i.e.$, $N$).
Additionally, we replace dense reconstructed graphons with sparse adjacency matrices of the sampled graphs with the help of the Bernoulli sampling. 
Therefore, it is relatively easy to compute the $d_{\text{fgw}}(\hat{\bm{x}}_{G_i},\bm{x}_n)$ in Line 7 of Algorithm~\ref{alg:raml} --- its computational complexity is $\mathcal{O}(EK)$ and $E$ is the number of the edges of the graph inducing $\bm{x}_n$. 
Moreover, the gradient corresponding to the first term of $\mathcal{L}$ is $-\sum_{n=1}^{N_b}\sum_{i=1}^{I}q(G_i|\bm{x}_n)\nabla\log p(G_i|h(\bm{z}_n))$. 
Here, the $q(G_i|\bm{x}_n)$ is used as a constant, and the corresponding FGW distance is not involved in the backpropagation, which reduces the cost of time and memory greatly.

\section{Connections to Existing Work}
\textbf{Autoencoders}
The principle of autoencoders is to minimize the discrepancy between the data and model distributions. 
The variational autoencoder (VAE)~\cite{kingma2013auto} and its variants~\cite{tomczak2018vae,wang2019topic} apply the KL-divergence as the discrepancy and learns a probabilistic autoencoder via maximizing the evidence lower bound (ELBO). 
The Wasserstein autoencoders (WAEs)~\cite{tolstikhin2018wasserstein,kolouri2018sliced} minimize a relaxed form of the Wasserstein distance to learn a deterministic autoencoder. 
Both these two strategies lead to a learning task including a reconstruction loss of observed data and a regularizer penalizing the distance between the prior and the posterior (or the mixture of different posterior distributions) in the latent space. 
The prior can be a predefined normal distribution or a learnable mixture model~\cite{takahashi2019variational,xu2020learning}. 
The commonly-used distances between the prior and the posterior include KL divergence, maximum mean discrepancy, GAN-based loss, FGW distance, $etc$.

\textbf{Generative graph modeling}
The early graph models like the Erd{\H{o}}s-R{\'e}nyi graph~\cite{erdHos1960evolution} simulate large graphs to yield certain statistical properties but cannot capture complicated mechanisms of real-world graphs. 
Recently, the GNN-based graph generative models have been widely used, which can be categorized into two classes. 
The first class learns node-level embeddings and estimates edges based on the pairs of the embeddings~\cite{kipf2016variational,kipf2017semi,niepert2016learning,xu2018powerful}, which works well on link prediction~\cite{zhang2018link} and conditional graph generation~\cite{yang2019conditional}. 
The second class applies various pooling layers~\cite{ying2018hierarchical,vinyals2016order,li2016gated} to obtain graph embeddings and then leverages recurrent neural networks to generate nodes and edges in an autoregressive manner~\cite{you2018graph,shi2019graphaf,jin2020hierarchical,dai2020scalable}.
Besides the GNN-based models, the Gromov-Wasserstein factorization (GWF) model~\cite{xu2020gromov} reconstructs each graph as a weighted GW barycenter~\cite{peyre2016gromov} of learnable graph factors, which achieves encouraging performance on graph clustering. 
Following the GWF model, the graph dictionary learning (GDL) model~\cite{vincent2021online} leverages a linear factorization to reconstruct graphs, which has lower complexity than the GWF model. 
However, the models above seldom consider the clustering structure or the distribution of the graph embeddings they learned. 

\textbf{Graphon-based graph models}
Graphon is a nonparametric graph model, which has been widely used in network modeling~\cite{avella2018centrality,gao2019graphon} and optimization~\cite{parise2018graphon}. 
To infer graphons from observed graphs, many methods have been proposed, $i.e.$, the stochastic block approximation (SBA) methods~\cite{airoldi2013stochastic,channarond2012classification,chan2014consistent} and the low-rank approximation methods~\cite{keshavan2010matrix,chatterjee2015matrix,xu2018rates}. 
These methods require well-aligned graphs, which is questionable in practice.
The work in~\cite{xu2021learning} relaxes this requirement, learning the graphon and aligning the observed graphs alternately by solving a GW barycenter problem~\cite{peyre2016gromov}. 
All the methods above are based on the weak regularity lemma~\cite{lovasz2012large}, approximating graphons by 2D step functions. 
Recently, the work in~\cite{ruiz2020graphon,ruiz2020graphon2} bridges the gap between graphon-based signal processing and graph neural networks, which inspires the design of our encoder.
However, existing methods either learn graphons to generate graphs or leverage graphons to process the information of nodes. 
None of them consider learning the distribution of graphons as we did.

\textbf{The novelties of our GNAE} To our knowledge, our graphon autoencoder makes the first attempt to build a WAE in the graphon space, which provides a new algorithmic framework for graphon distribution modeling and graph generation.
Our GNAE extends the GNN-based model and the factorization model to the functional space of graphons. 
The encoder achieves graphon filtering, whose GNN-based implementation is a special case for induced graphons. 
The decoder improves the GDL model by leveraging graphon factors with different partitions. 
Combining these two strategies in the framework of autoencoders, our GNAE inherits their advantages and learns them with better interpretability and capability.

\section{Experiments}\label{sec:exp}
\subsection{Graph representation and classification}
To demonstrate the usefulness of our GNAE model, we test it on six public graph datasets and compare it with state-of-the-art methods on graph modeling. 
The datasets we used can be categorized into three classes:
The \textbf{MUTAG} and the \textbf{PTC-MR} in~\cite{kriege2012subgraph} contain molecules with categorical node attributes;
the \textbf{PROTEIN} and the \textbf{ENZYMES} in~\cite{borgwardt2005protein} contain proteins with continuous node attributes;
and the \textbf{IMDB-B} and the \textbf{IMDB-M} in~\cite{yanardag2015deep} contain social networks without node attributes. 
These datasets can be downloaded from \url{https://chrsmrrs.github.io/datasets/}~\cite{Morris+2020}. 
For the datasets without node attributes, we treat the local degree profiles~\cite{cai2018simple} of nodes as the attributes.

The baselines include: ($i$) the kernel-based methods, $e.g.$, Random Walk Kernel (\textbf{RWK})~\cite{gartner2003graph}, Shortest Path Kernel (\textbf{SPK})~\cite{borgwardt2005protein}, Graphlet Kernel (\textbf{GK})~\cite{shervashidze2009efficient}, Weisfeiler-Lehman Sub-tree Kernel (\textbf{WLK})~\cite{shervashidze2011weisfeiler}, Deep Graph Kernel (\textbf{DGK})~\cite{yanardag2015deep}, Multi-Scale Laplacian Kernel (\textbf{MLGK})~\cite{kondor2016multiscale}, and Fused Gromov-Wasserstein Kernel (\textbf{FGWK})~\cite{titouan2019optimal}; 
($ii$) the GNN-based methods, $e.g.$, \textbf{sub2vec}~\cite{adhikari2018sub2vec}, \textbf{graph2vec}~\cite{narayanan2017graph2vec}, and the state-of-the-art InfoGraph method~\cite{sun2019infograph} that uses Graph Isomorphismic Network (GIN)~\cite{xu2018powerful} and Differentiable Pooling (DP)~\cite{ying2018hierarchical} as its backbone model, respectively (\textbf{InfoGraph$_{\text{GIN}}$} and \textbf{InfoGraph$_{\text{DP}}$});
($iii$) the factorization models (FMs), $e.g.$, the Gromov-Wasserstein factorization (\textbf{GWF})~\cite{xu2020gromov} and the Graph Dictionary Learning (\textbf{GDL})~\cite{vincent2021online}.
We reproduce the baselines either based on the code released by the authors or our own implementations and set their hyperparameters according to the released code or the corresponding references. 
When implementing our GNAE model, we consider two variants: applying the FGW distance directly as the underlying distance and learning the GNAE model by alternating optimization (\textbf{GNAE}$_{\textbf{FGW}}$), or applying the KL divergence of graph distributions as the underlying distance and learning the model by the proposed RAML (\textbf{GNAE}$_{\textbf{RAML}}$). 
For the GNAE models, the settings of their hyperparameters are given in Appendix.

We test the methods above on graph classification. 
For each kernel-based method, we train a kernel SVM classifier~\cite{chang2011libsvm}.
For other methods, we learn graph representations explicitly in an unsupervised way and train an SVM classifier based on the representations. 
The SVM classifier of each method is trained based on 10-fold cross-validation, and we use the same random seed to split data and select the most suitable SVM kernel function manually. 
Table~\ref{tab:class} lists the mean and the standard deviation of the classification accuracy achieved by the methods on each dataset. 
We can find that the performance of our GNAE models is at least comparable to that of the state-of-the-art methods ($e.g.$, MLGK, FGWK and InfoGraph). 
Especially, the proposed GNAE$_{\text{RAML}}$ achieves the top-5 accuracy on four of the six datasets, which is the same as InforGraph$_{\text{GIN}}$ does. 
Note that for the GNN-based methods, the dimension of their graph representations is over $100$. 
However, our GNAE models achieve competitive results based on the representations with much a lower dimension ($\leq 30$ for all the datasets).
For the challenging ENZYMES dataset, our GNAE methods do not work well. 
A potential reason for this phenomenon is the model misspecification issue --- the node attributes in this dataset are sparse and have high dynamic ranges, so the smoothed signal model we applied may not be able to describe and reconstruct such attributes well. 

\begin{table}[!t]
\centering
\caption{Comparison on classification accuracy ($\%$).}
\label{tab:class}
\begin{small}
\begin{threeparttable}
\begin{tabular}{
    @{\hspace{1pt}}c@{\hspace{2pt}}|
	@{\hspace{2pt}}l@{\hspace{3pt}}|
	@{\hspace{3pt}}c@{\hspace{3pt}}
	@{\hspace{3pt}}c@{\hspace{3pt}}
	@{\hspace{3pt}}c@{\hspace{3pt}}
	@{\hspace{3pt}}c@{\hspace{3pt}}
	@{\hspace{3pt}}c@{\hspace{3pt}}
	@{\hspace{3pt}}c@{\hspace{3pt}}|
	@{\hspace{3pt}}c@{\hspace{1pt}}
}
\hline\hline
Category
&Method
&MUTAG 
&PTC-MR 
&PROTEIN
&ENZYMES
&IMDB-B 
&IMDB-M
&\# in Top5\\ \hline
\multirow{6}{*}{Kernels}
&RWK                               
&83.72$_{\pm \text{1.50}}$                  
&57.85$_{\pm \text{1.30}}$   
&73.95$_{\pm \text{0.59}}$
&28.52$_{\pm \text{1.83}}$
&50.70$_{\pm \text{0.26}}$                  
&34.65$_{\pm \text{0.19}}$
&0\\ 
&SPK                                    
&\textbf{85.22}$_{\pm \text{2.43}}$                  
&58.24$_{\pm \text{2.44}}$   
&\textbf{74.93}$_{\pm \text{0.86}}$
&38.87$_{\pm \text{3.01}}$
&55.60$_{\pm \text{0.22}}$                  
&37.99$_{\pm \text{0.30}}$
&2\\
&GK                                     
&81.66$_{\pm \text{2.11}}$                  
&57.26$_{\pm \text{1.41}}$    
&71.10$_{\pm \text{1.08}}$
&30.36$_{\pm \text{4.84}}$
&65.90$_{\pm \text{0.98}}$                  
&43.89$_{\pm \text{0.38}}$
&0\\
&WLK                                   
&80.72$_{\pm \text{3.00}}$                  
&57.97$_{\pm \text{0.49}}$    
&73.01$_{\pm \text{1.09}}$
&54.69$_{\pm \text{3.27}}$
&\textbf{72.30}$_{\pm \text{3.44}}$                  
&46.35$_{\pm \text{0.46}}$
&1\\ 
&DGK                                    
&\textbf{87.44}$_{\pm \text{2.72}}$ 
&60.08$_{\pm \text{2.55}}$
&74.27$_{\pm \text{1.12}}$
&53.22$_{\pm \text{1.01}}$
&66.90$_{\pm \text{0.56}}$ 
&44.55$_{\pm \text{0.52}}$
&1\\ 
&MLGK
&\textbf{87.94}$_{\pm \text{1.61}}$ 
&\textbf{62.23}$_{\pm \text{1.39}}$
&\textbf{75.86}$_{\pm \text{0.99}}$
&61.89$_{\pm \text{1.17}}$
&66.60$_{\pm \text{0.25}}$ 
&41.17$_{\pm \text{0.03}}$
&3\\ 
&FGWK\tnote{*}
&\textbf{88.13}$_{\pm \text{4.22}}$ 
&\textbf{62.98}$_{\pm \text{5.27}}$
&72.20$_{\pm \text{3.81}}$
&\textbf{71.48}$_{\pm \text{2.96}}$
&63.50$_{\pm \text{4.01}}$ 
&46.27$_{\pm \text{3.85}}$
&3\\ \hline
\multirow{4}{*}{GNNs}
&sub2vec
&60.88$_{\pm \text{9.89}}$                 
&59.99$_{\pm \text{6.38}}$  
&54.29$_{\pm \text{5.20}}$
&45.25$_{\pm \text{2.80}}$
&55.30$_{\pm \text{1.54}}$                  
&36.67$_{\pm \text{0.83}}$
&0\\ 
&graph2vec                         
&83.15$_{\pm \text{9.25}}$                  
&60.17$_{\pm \text{6.86}}$   
&72.96$_{\pm \text{1.89}}$
&\textbf{71.65}$_{\pm \text{3.10}}$
&71.10$_{\pm \text{0.54}}$              
&\textbf{50.44}$_{\pm \text{0.87}}$
&2\\
&InfoGraph$_{\text{GIN}}$                      
&\textbf{89.13}$_{\pm \text{1.01}}$
&61.65$_{\pm \text{1.43}}$
&\textbf{74.88}$_{\pm \text{4.31}}$
&39.52$_{\pm \text{3.99}}$
&\textbf{73.90}$_{\pm \text{0.87}}$         
&\textbf{49.29}$_{\pm \text{0.53}}$
&\textbf{4}\\
&InfoGraph$_{\text{DP}}$\tnote{*}                  
&84.28$_{\pm \text{3.94}}$        
&\textbf{62.26}$_{\pm \text{4.55}}$
&73.50$_{\pm \text{2.91}}$
&\textbf{61.93}$_{\pm \text{4.64}}$
&68.50$_{\pm \text{5.07}}$         
&44.79$_{\pm \text{3.33}}$
&2\\\hline
\multirow{2}{*}{FMs}
&GWF
&78.25$_{\pm \text{3.67}}$    
&\textbf{61.87}$_{\pm \text{2.53}}$  
&73.19$_{\pm \text{1.97}}$
&\textbf{72.11}$_{\pm \text{4.00}}$
&60.90$_{\pm \text{2.68}}$    
&39.97$_{\pm \text{1.35}}$
&2\\
&GDL\tnote{*}
&78.18$_{\pm \text{2.37}}$ 
&60.32$_{\pm \text{1.35}}$  
&74.29$_{\pm \text{3.60}}$
&\textbf{71.15}$_{\pm \text{3.19}}$
&\textbf{71.70}$_{\pm \text{1.10}}$ 
&\textbf{49.12}$_{\pm \text{0.49}}$ 
&3\\ \hline
\multirow{2}{*}{\textbf{Ours}}
&GNAE$_{\text{FGW}}$
&79.53$_{\pm\text{5.79}}$
&61.43$_{\pm\text{4.28}}$
&\textbf{75.32}$_{\pm\text{2.88}}$
&48.00$_{\pm\text{6.36}}$
&\textbf{72.50}$_{\pm\text{4.30}}$
&\textbf{47.30}$_{\pm\text{1.97}}$ 
&3\\ 
&GNAE$_{\text{RAML}}$
&79.76$_{\pm\text{3.88}}$   
&\textbf{61.75}$_{\pm\text{6.29}}$
&\textbf{75.78}$_{\pm\text{3.42}}$
&50.70$_{\pm\text{4.14}}$
&\textbf{73.10}$_{\pm\text{3.75}}$
&\textbf{46.67}$_{\pm\text{3.33}}$
&\textbf{4}\\ 
\hline\hline
\end{tabular}
\begin{footnotesize}
\begin{tablenotes}
    \item[1] The methods marked by ``*'' are implemented by ourselves.
    \item[2] For each dataset, the bold numbers are the five highest accuracy (top-5 results).
\end{tablenotes}
\end{footnotesize}
\end{threeparttable}
\end{small}
\end{table}

\begin{figure}[t]
    \centering
    \subfigure[Training time]{
    \includegraphics[height=2.7cm]{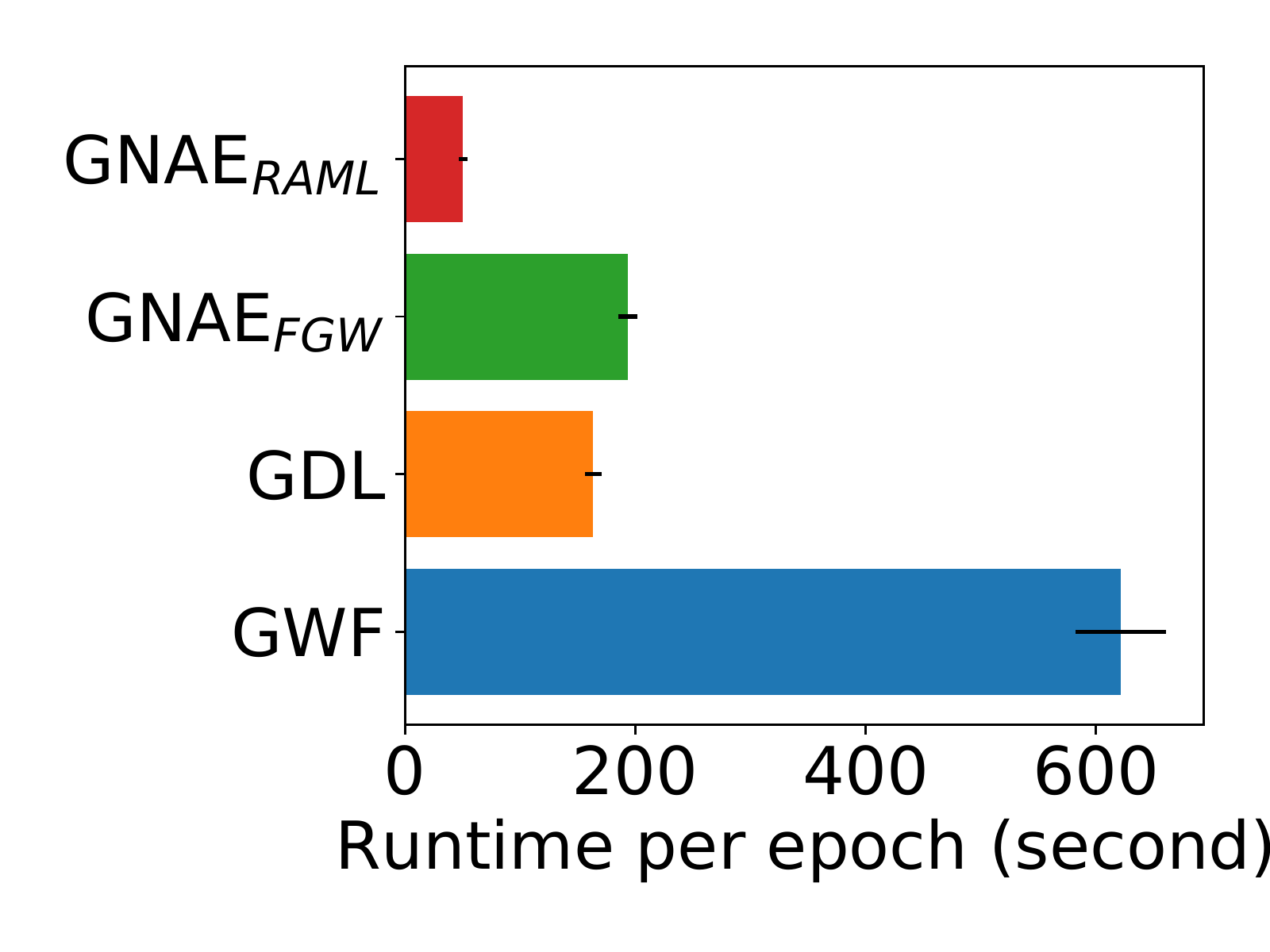}\label{fig:time}
    }
    \subfigure[Typical graphs and graphon factors]{
    \includegraphics[height=2.7cm]{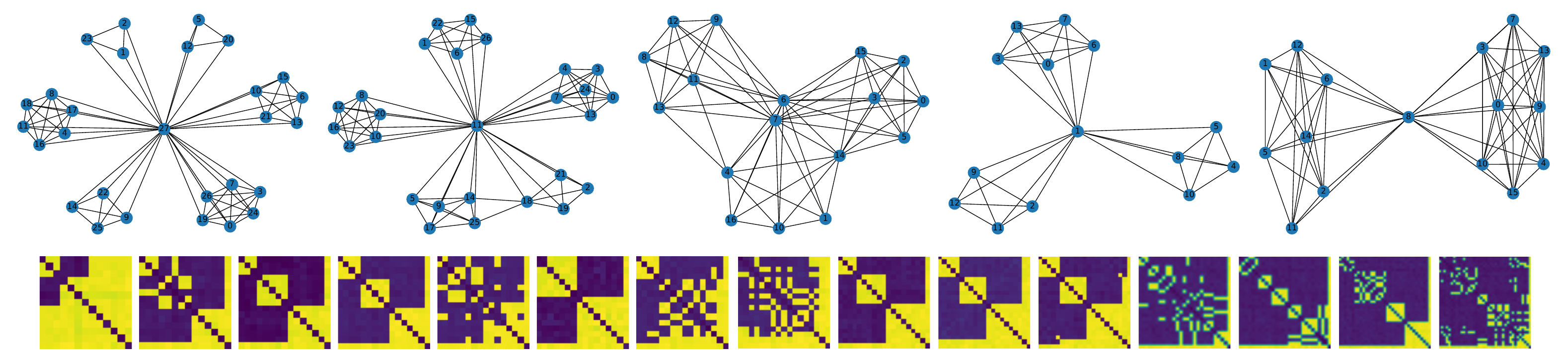}\label{fig:factors}
    }
    \caption{For the IMDB-B dataset: (a) Comparisons for FGW-based methods on their runtime. (b) Illustrations of typical graphs and the graphon factors learned by our GNAE$_{\text{RAML}}$.}
\end{figure}

Our GNAE$_{\text{RAML}}$ is more efficient than its competitors that apply FGW distance ($i.e.$, GWF, GDL, and GNAE$_{\text{FGW}}$). 
Suppose that all the models contain $C$ graph (or graphon) factors with comparable sizes, denoted as $\mathcal{O}(N)$. 
Given a graph with $N$ nodes and $E$ edges, the GWF reconstructs it by the FGW barycenter of its factors~\cite{xu2020gromov}, which needs to compute $C$ FGW distances iteratively. 
Therefore, its computational complexity is at least $\mathcal{O}(CN^3)$. 
Both the GNAE$_{\text{FGW}}$ and the GDL applies linear factorization models, so they only need to compute one FGW distance between the input and the reconstruction, whose complexity is $\mathcal{O}(|\mathcal{P}|^2N)$ and $\mathcal{O}(N^3)$, respectively. 
Here, $\mathcal{P}$ is the partitions of the graphon reconstructed by the GNAE$_{\text{FGW}}$. 
Proposition~\ref{prop:partition} shows that $|\mathcal{P}|\geq N$, so the GNAE$_{\text{FGW}}$ is slightly slower than the GDL.
Our GNAE$_{\text{RAML}}$ samples $I$ small graphs and computes $I$ FGW distances, each of which is a pair of two sparse matrices. 
Denote the number of nodes in each small graph as $K$. 
The computational complexity of our GNAE$_{\text{RAML}}$ is $\mathcal{O}(IEK)$.
Because $E\ll N^2$, $K\ll N$, and we set $I=\mathcal{O}(C)$, our GNAE$_{\text{RAML}}$ owns the lowest computational complexity. 
Figure~\ref{fig:time} shows the training time per epoch of different models on the IMDB-B dataset, which verifies our analysis above --- our GNAE$_{\text{RAML}}$ is $\times 4$ faster than the GDL and GNAE$_{\text{FGW}}$ and $\times 12$ faster than the GWF. 
Note that because the implementation of the GDL does not support GPU computing, we test all the methods on a single core of a CPU (Core i7 2.5GHz) for fairness.
Figure~\ref{fig:factors} visualize some typical graphs in the IMDB-B dataset and the graphon factors learned by our GNAE$_{\text{RAML}}$ ($i.e.$, $\{\tilde{g}_c\}_{c=1}^{15}$). 
We can find that the IMDB-B graphs are formulated as communities connected by one or two central nodes. 
The graphon factors we learned reflect the topological property of the graphs, which further demonstrates the rationality of our GNAE model.

\subsection{Generalizability and transferability on social network modeling}
\begin{figure}[t]
    \centering
    \includegraphics[height=2.5cm]{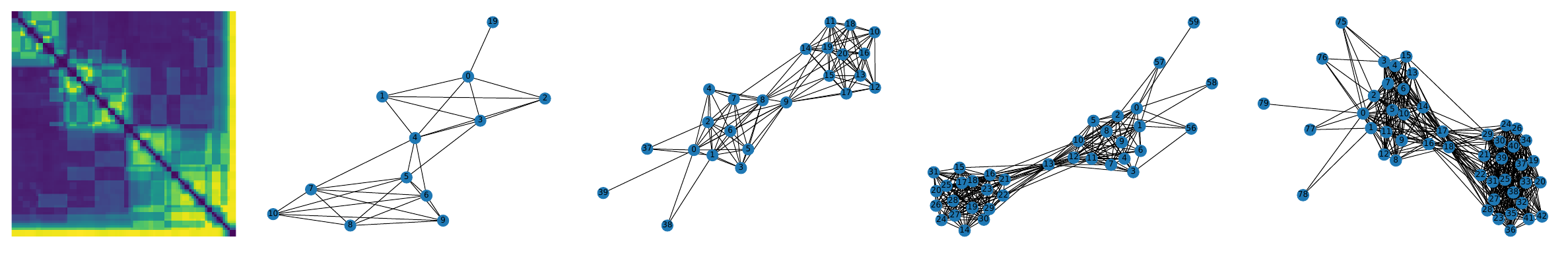}
    \caption{Illustrations of the graphs sampled from the generated graphon on the left. From left to right, the number of nodes for each graph is $20, 40, 60, 80$, respectively.}
    \label{fig:generate}
\end{figure}

Our GNAE can generate graphons from graph representations and sampling graphs with different sizes but similar topological structures. 
Again, take the IMDB-B dataset as an example. 
For this dataset, the average number of nodes per graph is 19.77. 
Given a GNAE trained on this dataset, we sample graph representations from learned prior distribution and generate graphons by the decoder of the GNAE, $i.e.$, $\hat{g}=h(\bm{z})$ with $\bm{z}\sim p_{\mathcal{Z}}$. 
Based on $\hat{g}$, we sample graphs with different sizes, as shown in Figure~\ref{fig:generate}.
We can find that the generated graphs have similar structures, each containing two communities connected by few key nodes.
Note that this topological structure is typical for the real IMDB graph (as shown in Figure~\ref{fig:factors}).
This experimental result demonstrates that our GNAE has the potentials as a graph generator with strong generalizability, which is especially suitable for social network modeling and simulation. 

Another advantage of our GNAE is its transferability, which is seldom considered by existing work. 
In particular, we can train a GNAE on a dataset and use it to represent the graphs in a related but different dataset. 
For example, both the IMDB-B and the IMDB-M are movie collaboration datasets. 
Each graph in these two datasets is an ego-network of an actor/actress, which indicates his/her collaborations with other actors/actresses~\cite{yanardag2015deep}.   
The IMDB-B contains 1000 ego-networks driven by two genres (\textit{Action} and \textit{Romance}), while the IMDB-M contains 1500 ego-networks driven by three genres (\textit{Comedy}, \textit{Romance} and \textit{Sci-Fi}). 
Obviously, these two datasets have different structures but share some information. 
To demonstrate the transferability of our model, we first train a GNAE model on one dataset. 
Then, without any fine-tuning, we leverage the model to represent the graphs in the other dataset. 
Finally, we train and test an SVM classifier based on the representations and record the classification accuracy achieved by 10-fold cross-validation.
Table~\ref{tab:trans} shows the performance of our GNAE and the strongest baseline InfoGraph$_{\text{GIN}}$ in the transfer learning scenarios. 
We can find that the performance of the InfoGraph$_{\text{GIN}}$ drops a lot when doing transfer learning. 
On the contrary, our GNAE shows good transferability, whose performance only degrades slightly. 
It captures the structural information shared by the two datasets, making the model transferable.

\begin{table}[t]
    \centering
    \caption{Comparison on the classification accuracy ($\%$) achieved by transfer learning}
    \begin{small}
    \begin{tabular}{
    @{\hspace{1pt}}c|
    c@{\hspace{3pt}}|
    @{\hspace{3pt}}c@{\hspace{3pt}}|
    @{\hspace{3pt}}c@{\hspace{3pt}}|
    @{\hspace{3pt}}c@{\hspace{1pt}}}
    \hline\hline
    \multirow{2}{*}{Method} & \multicolumn{4}{c}{Training $\rightarrow$ Testing}\\
    \cline{2-5}
    &
    IMDB-B $\rightarrow$ IMDB-B &
    IMDB-M $\rightarrow$ IMDB-B &
    IMDB-M $\rightarrow$ IMDB-M &
    IMDB-B $\rightarrow$ IMDB-M \\
    \hline
    InfoGraph$_{\text{GIN}}$ &
    73.90$_{\pm \text{0.87}}$ &
    66.10$_{\pm \text{1.90}}$ &
    49.29$_{\pm \text{0.53}}$ &
    45.29$_{\pm \text{1.28}}$\\
    GNAE$_{\text{RAML}}$ &
    73.60$_{\pm \text{3.80}}$ &
    70.70$_{\pm \text{3.49}}$ &
    46.93$_{\pm \text{3.14}}$ &
    46.20$_{\pm \text{3.50}}$\\
    \hline\hline
    \end{tabular}
    \end{small}
    \label{tab:trans}
\end{table}

\section{Conclusion and Future Work}\label{sec:conclusion}
We proposed a novel graphon autoencoder associated with an efficient learning algorithm.
It is pioneering work achieving an interpretable and scalable graph generative model. 
Currently, the main advantages of our GNAE, $e.g.$, its generalizability and transferability, are demonstrated on social network modeling. 
However, as shown in Table~\ref{tab:class}, we need to improve the GNAE model for other graph types like proteins, molecules, and more complicated heterogeneous graphs and hypergraphs. 
Additionally, we will explore other potential substitutes for the FGW distance to further improve the efficiency of our learning algorithm.

\bibliography{gnae}
\bibliographystyle{plain}

\newpage

\section{Appendix}
\subsection{The derivation of (\ref{eq:d_fgw2})}
Suppose that $\mathcal{P}=\{\mathcal{P}_n\}_{n=1}^{N}$ and $\mathcal{Q}=\{\mathcal{Q}_m\}_{m=1}^{M}$ are two sets of partitions in $\Omega$.
Given $\bm{x}_{1,\mathcal{P}}=(g_{1,\mathcal{P}},s_{1,\mathcal{P}})$ and $\bm{x}_{2,\mathcal{Q}}=(g_{2,\mathcal{Q}},s_{2,\mathcal{Q}})$, where $g_{1,\mathcal{P}}(v,v')=\sum_{n,n'=1}^{N}g_{1,nn'}1_{\mathcal{P}_n}(v)1_{\mathcal{P}_{n'}}(v')$, $g_{2,\mathcal{Q}}(v,v')=\sum_{m,m'=1}^{M}g_{2,mm'}1_{\mathcal{Q}_m}(v)1_{\mathcal{Q}_{m'}}(v')$, $s_{1,\mathcal{P}}(v)=\sum_{n=1}^{N}\bm{s}_{1,n} 1_{\mathcal{P}}(v)$, and $s_{2,\mathcal{Q}}(v)=\sum_{m=1}^{M}\bm{s}_{2,m} 1_{\mathcal{Q}}(v)$ are step functions, we have
\begin{eqnarray*}
\begin{aligned}
&d_{\text{fgw}}(\bm{x}_{1,\mathcal{P}}, \bm{x}_{2,\mathcal{Q}})\\
=&\inf_{\pi\in\Pi(\mu_{\Omega},\mu_{\Omega})}\Bigl(\int_{\Omega^2\times\Omega^2}|g_{1,\mathcal{P}}(u,u')-g_{2,\mathcal{Q}}(v,v')|^p d\pi(u,v)d\pi(u',v')+\\
&\int_{\Omega^2}\|s_{1,\mathcal{P}}(u)-s_{2,\mathcal{Q}}(v)\|_p^p d\pi(u,v)\Bigr)^{\frac{1}{p}}\\
=&\inf_{\pi\in\Pi(\mu_{\Omega},\mu_{\Omega})}\Bigl(\int_{\Omega^2\times\Omega^2}|\sum_{n,n'}g_{1,nn'}1_{\mathcal{P}_n}(u)1_{\mathcal{P}_{n'}}(u')-\\
&\sum_{m,m'}g_{2,mm'}1_{\mathcal{Q}_m}(v)1_{\mathcal{Q}_{m'}}(v')|^p d\pi(u,v)d\pi(u',v')+\\
&\int_{\Omega^2}\|\sum_{n}\bm{s}_{1,n}1_{\mathcal{P}_m}(u)-\sum_{m}\bm{s}_{2,m}1_{\mathcal{Q}_m}(v)\|_p^p d\pi(u,v)\Bigr)^{\frac{1}{p}}\\
=&\inf_{\pi\in\Pi(\mu_{\Omega},\mu_{\Omega})}\Bigl(\sum_{n,n',m,m'}\int_{\mathcal{P}_n\times\mathcal{P}_{n'}\times\mathcal{Q}_m\times\mathcal{Q}_{m'} }|g_{1,nn'}-g_{2,mm'}|^p d\pi(u,v)d\pi(u',v')+\\
&\sum_{n,m}\int_{\mathcal{P}_{n}\times\mathcal{Q}_m}\|\bm{s}_{1,n}-\bm{s}_{2,m}\|_p^p d\pi(u,v)\Bigr)^{\frac{1}{p}}\\
=&\inf_{\pi\in\Pi(\mu_{\Omega},\mu_{\Omega})}\Bigl(\sum_{n,n',m,m'}|g_{1,nn'}-g_{2,mm'}|^p\underbrace{\int_{\mathcal{P}_n\times\mathcal{Q}_m} d\pi(u,v)}_{t_{nm}}\underbrace{\int_{\mathcal{P}_{n'}\times\mathcal{Q}_{m'}}d\pi(u',v')}_{t_{n'm'}}+\\
&\sum_{n,m}\|\bm{s}_{1,n}-\bm{s}_{2,m}\|_p^p\underbrace{\int_{\mathcal{P}_{n}\times\mathcal{Q}_m} d\pi(u,v)}_{t_{nm}}\Bigr)^{\frac{1}{p}}\\
=&\sideset{}{_{\bm{T}\in\Pi(\bm{\mu}_{\mathcal{P}},\bm{\mu}_{\mathcal{Q}})}}\min ( \langle\bm{D}_g, \bm{T}\otimes\bm{T}\rangle + \langle\bm{D}_s, \bm{T}\rangle )^{\frac{1}{p}},
\end{aligned}
\end{eqnarray*}
where $\bm{D}_g=[|g_{1,nn'}-g_{2,mm'}|^p]\in\mathbb{R}^{N^2\times M^2}$, $\bm{D}_s=[\|\bm{s}_{1,n}-\bm{s}_{2,m}\|_p^p]\in\mathbb{R}^{N\times M}$, $\otimes$ represents Kronecker product, $\bm{T}=[t_{nm}]$, and it is in the set $\Pi(\bm{\mu}_{\mathcal{P}},\bm{\mu}_{\mathcal{Q}})=\{\bm{T}\geq\bm{0}|\bm{T}\bm{1}=\bm{\mu}_{\mathcal{P}},\bm{T}^{\top}\bm{1}=\bm{\mu}_{\mathcal{Q}}\}$ with $\bm{\mu}_{\mathcal{P}}=[\frac{|\mathcal{P}_1|}{|\Omega|},..,\frac{|\mathcal{P}_N|}{|\Omega|}]$ and $\bm{\mu}_{\mathcal{Q}}=[\frac{|\mathcal{Q}_1|}{|\Omega|},..,\frac{|\mathcal{Q}_M|}{|\Omega|}]$.

\subsection{The computation of the FGW distance}
For the FGW distance shown in (\ref{eq:d_fgw2}), we set $p=2$ as~\cite{peyre2016gromov}.  
Denote $\bm{G}_1=[g_{1,nn'}]$ and $\bm{G}_2=[g_{2,mm'}]$ as the matrices corresponding to the step functions $g_{1,\mathcal{P}}$ and $g_{2,\mathcal{Q}}$. 
Applying $p=2$, we can rewrite the objective function as
\begin{eqnarray}
\langle \bm{D}(\bm{T}),\bm{T}\rangle = \langle \bm{D}_s + \bm{G}_{12}-2\bm{G}_1\bm{T}\bm{G}_2^{\top},\bm{T}\rangle,
\end{eqnarray}
where $\bm{G}_{12}=(\bm{G}_1\odot\bm{G}_1)\bm{\mu}_{\mathcal{P}}\bm{1}_{M}^{\top} + \bm{1}_N\bm{\mu}_{\mathcal{Q}}^{\top}(\bm{G}_2\odot\bm{G}_2)^{\top}$~\cite{peyre2016gromov}. 
$\odot$ represents Hadamard product and $\bm{1}_N$ is $N$-dimensional all-one vector. 

We apply the proximal gradient algorithm in~\cite{xu2020gromov} to compute the FGW distance, which is shown in Algorithm~\ref{alg:sinkhorn}.
This algorithm ensures the optimal transport matrix to converge to a stationary point. 
In this work, we set the number of iterations $J_1=20$ and the number of Sinkhorn iterations $J_2=5$ when computing FGW distance.
\begin{algorithm}[ht]
    \caption{$\min_{\bm{T}\in\Pi(\bm{\mu}_{\mathcal{P}},\bm{\mu}_{\mathcal{Q}})}\langle\bm{D}_g, \bm{T}\otimes\bm{T}\rangle + \langle\bm{D}_s, \bm{T}\rangle$}	
    \label{alg:sinkhorn}
	\begin{algorithmic}[1]
	    \STATE Compute $\bm{G}_{12}=(\bm{G}_1\odot\bm{G}_1)\bm{\mu}_{\mathcal{P}}\bm{1}_{M}^{\top} + \bm{1}_N\bm{\mu}_{\mathcal{Q}}^{\top}(\bm{G}_2\odot\bm{G}_2)^{\top}$
	    \STATE Initialize $\bm{T}^{(0)}=\bm{\mu}_{\mathcal{P}}\bm{\mu}_{\mathcal{Q}}^{\top}$, $\bm{a}=\bm{\mu}_{\mathcal{P}}$.
		\STATE \textbf{for} $j=0,...,J_1-1$ 
		\STATE \quad $\bm{C}=\exp(-\frac{1}{\beta}(\bm{D}_s+\bm{G}_{12}-2\bm{G}_1\bm{T}\bm{G}_2^{\top}))\odot \bm{T}^{(j)}$
		\STATE \quad Sinkhorn iteration: \textbf{for} $n=0,..,J_2-1$,  $\bm{b}=\frac{\bm{\mu}_{\mathcal{Q}}}{\bm{C}^{\top}\bm{a}}$,  $\bm{a} = \frac{\bm{\mu}_{\mathcal{P}}}{\bm{C}\bm{b}}$,
		\STATE \quad $\bm{T}^{(j + 1)} = \text{diag}(\bm{a})\bm{C}\text{diag}(\bm{b})$.
		\STATE The optimal transport $\bm{T}^*=\bm{T}^{(J)}$.
	\end{algorithmic}
\end{algorithm}

\subsection{The computation of the sliced FGW distance}
\begin{definition}[Sliced FGW Distance]
Denote $\mathcal{S}^{M-1}=\{\bm{\theta}\in\mathbb{R}^M | \|\bm{\theta}\|_2=1\}$ as the $M$-dimensional hypersphere and $\mu_{\mathcal{S}^{M-1}}$ the probability measure on $\mathcal{S}^{M-1}$. 
For the probability measures $p_\mathcal{Z}$ and $q_{\mathcal{Z}}$ on the metric space $(\mathcal{Z}, d_\mathcal{Z})$, their sliced fused Gromov-Wasserstein distance is
\begin{eqnarray*}
d_{\text{sfgw}}(p_\mathcal{Z}, q_{\mathcal{Z}}) = \mathbb{E}_{\bm{\theta}\sim \mu_{\mathcal{S}^{M-1}}}[d_{\text{fgw}}(R_{\bm{\theta}}\#p_\mathcal{Z}, R_{\bm{\theta}}\#q_\mathcal{Z})],
\end{eqnarray*}
where $R_{\bm{\theta}}$ as the projection on $\bm{\theta}$, where $R_{\bm{\theta}}(\bm{z})=\langle \bm{z}, \bm{\theta}\rangle$. $R_{\bm{\theta}}\#p$ represents the distribution after the projection, and $d_{\text{fgw}}(R_{\bm{\theta}}\#p_\mathcal{Z}, R_{\bm{\theta}}\#q_\mathcal{Z})$ is the FGW distance between the $R_{\bm{\theta}}\#p_\mathcal{Z}$ and $R_{\bm{\theta}}\#q_\mathcal{Z}$ defined on the 1D metric space $(R_{\bm{\theta}}(\mathcal{Z}), d_{R_{\bm{\theta}}(\bm{Z})})$.
\end{definition}

In our case, we compute the order-2 sliced FGW distance between the expected posterior $q_{\mathcal{Z};f}$ and the prior $p_{\mathcal{Z}}$, $i.e.$, $d_{\text{sfgw}}(q_{\mathcal{Z};f},p_{\mathcal{Z}})$. 
The sliced FGW distance is the expectation of 1D FGW distances under different projections.
We can approximate it based on the samples of the distributions and the samples of the projections.
In particular, given $\{\bm{z}_{1,i}\}_{i=1}^{N}\sim q_{\mathcal{Z};f}$, $\{\bm{z}_{2,i}\}_{i=1}^{N}\sim p_{\mathcal{Z}}$, and $L$ projections $\{R_{\bm{\theta}_l}\}_{l=1}^{L}$, where $\bm{\theta}\sim \mu_{\mathcal{S}^{M-1}}$, the empirical sliced FGW, denoted as $\hat{d}_{\text{sfgw}}$, is
\begin{eqnarray*}\label{eq:esfgw}
\begin{aligned}
&\hat{d}_{\text{sfgw}}(q_{\mathcal{Z};f},p_{\mathcal{Z}}) \\
=&\frac{1}{L}\sideset{}{}\sum_{l=1}^{L} \hat{d}_{\text{fgw}}(R_{\bm{\theta}_l}\#q_{\mathcal{Z};f}, R_{\bm{\theta}_l}\#p_{\mathcal{Z}})\\
=&\frac{1}{NL}\sideset{}{}\sum_{l=1}^{L}\min_{\sigma\in \mathcal{U}_N}\sideset{}{}\sum_{i,j=1}^{N}((R_{\bm{\theta}_l}(\bm{z}_{1,i}) - R_{\bm{\theta}_l}(\bm{z}_{1,j}))^2 - (R_{\bm{\theta}_l}(\bm{z}_{2,\sigma(i)}) - R_{\bm{\theta}_l}(\bm{z}_{2,\sigma(j)}))^2)^2+\\
&\sideset{}{}\sum_{i=1}^{N}(R_{\bm{\theta}_l}(\bm{z}_{1,i}) - R_{\bm{\theta}_l}(\bm{z}_{2,\sigma(i)}))^2\\
=&\frac{1}{NL}\sideset{}{}\sum_{l=1}^{L}\min_{\sigma\in\{\sigma_a,\sigma_d\} }\sideset{}{}\sum_{i,j=1}^{N}((R_{\bm{\theta}_l}(\bm{z}_{1,\sigma_a(i)}) - R_{\bm{\theta}_l}(\bm{z}_{1,\sigma_a(j)}))^2 - (R_{\bm{\theta}_l}(\bm{z}_{2,\sigma(i)}) - R_{\bm{\theta}_l}(\bm{z}_{2,\sigma(j)}))^2)^2+\\
&\sideset{}{}\sum_{i=1}^{N}(R_{\bm{\theta}_l}(\bm{z}_{1,\sigma_a(i)}) - R_{\bm{\theta}_l}(\bm{z}_{2,\sigma(i)}))^2.
\end{aligned}
\end{eqnarray*}
Here, $\sigma(\cdot)\in\mathcal{U}_N$ represents a permutation of $\{1,...,N\}$. 
$\sigma_a(\cdot)$ ($\sigma_d(\cdot)$) outputs the indices of the samples in an ascending (descending) order.
The second equation is based on the fact that when computing the empirical FGW distance ($i.e.$, $\hat{d}_{\text{sfgw}}$) between a pair of $N$-sample sets in 1D space, the optimal transport matrix is a permutation matrix~\cite{xu2020learning}. 
The third equation is based on the Theorem 3.4 in~\cite{xu2020learning} --- the empirical FGW distance in 1D space corresponds to the distance between the samples in either identity or anti-identity order. 
In our work, the number of samples $N$ is equal to the batch size we set. The number of projections we used is $L=50$.

\subsection{The setting of hyperparameters}
We set the hyperparameters of our GNAE empirically.
Some hyperparameters are fixed for all the datasets: the batch size is $N_b=50$; the learning rate is $0.005$; the number of epochs is $25$; the number of sampled graphs per graphon is $I=5$; the size of the sampled graphs is $K=10$; the weight of regularizer $\gamma=0.1$; the number of ChebConv layers is $J=4$; the order of the FGW distance is $p=2$.

The other hyperparameters are specified for various datasets, $e.g.$, the number of Gaussian components in the prior distribution is equal to the number of clusters in each dataset. 
According to the type of signal (or equivalently, the type of node attribute), we set the distribution of signal in (\ref{eq:generate_graph},~\ref{eq:pG}) and the activation function $\alpha(\cdot)$ in (\ref{eq:reconstruct}), as shown in Table~\ref{tab:signal}.
The other settings are listed in Table~\ref{tab:hyper}.

\begin{table}[t]
    \centering
    \caption{The setting related to the type of signal}
    \label{tab:signal}
    \begin{small}
    \begin{threeparttable}
    \begin{tabular}{c|c|c|c}
    \hline\hline
    Signal type & 
    Signal distribution &
    $p(\bm{s}|s(v))$ & 
    $\alpha(\cdot)$\\
    \hline
    Continuous &
    Gaussian &
    $\propto \exp(-\|\bm{s}-s(v)\|_2^2/2M\sigma^2)$ &
    ---\\
    Binary &
    Bernoulli &
    $\prod_{m=1}^{M}s_m(v)^{\bm{s}_m}(1-s_m(v))^{1-\bm{s}_m}$ &
    sigmoid\\
    One hot &
    Categorical &
    $\prod_{m=1}^{M}s_m(v)^{\bm{s}_m}$ &
    softmax\\
    \hline\hline
    \end{tabular}
    \end{threeparttable}
    \end{small}
\end{table}

\begin{table}[t]
    \centering
    \caption{The setting of other hyperparameters}
    \label{tab:hyper}
    \begin{small}
    \begin{tabular}{c|cccccc}
    \hline\hline
    Dataset &  
    MUTAG & 
    PTC-MR &
    PROTEIN &
    ENZYMES &
    IMDB-B &
    IMDB-M \\
    \hline
    $J$ & 
    8 &
    4 &
    4 &
    8 &
    4 &
    4 \\
    $D$ &
    30 &
    30 &
    30 &
    50 &
    30 &
    30 \\
    $C$ &
    15 &
    15 &
    30 &
    30 &
    15 &
    15 \\
    \hline\hline
    \end{tabular}
    \end{small}
\end{table}


\begin{figure}[t]
    \centering
    \subfigure[MUTAG]{
    \includegraphics[height=6cm]{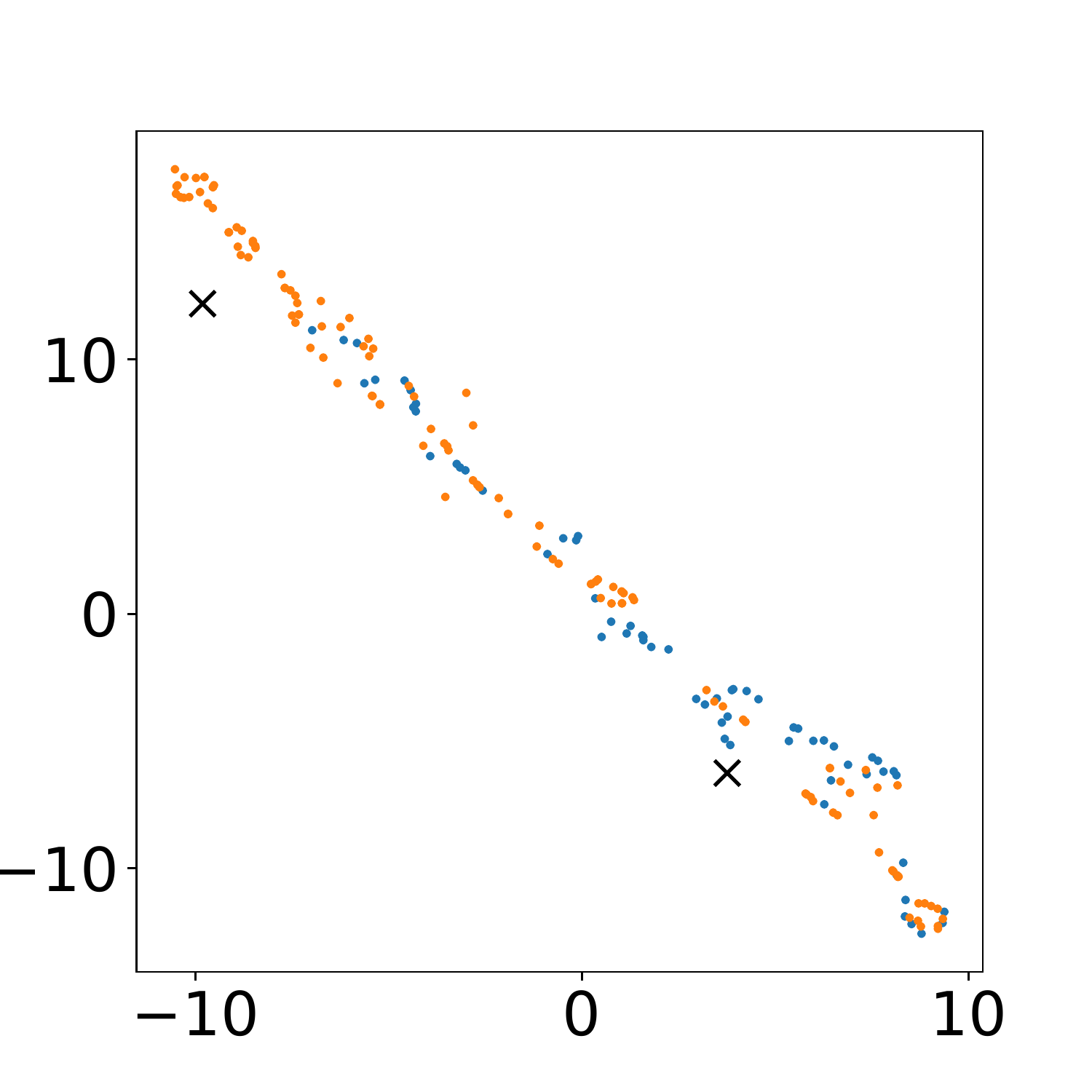}
    }
    \hspace{-3mm}
    \subfigure[AIDS]{
    \includegraphics[height=6cm]{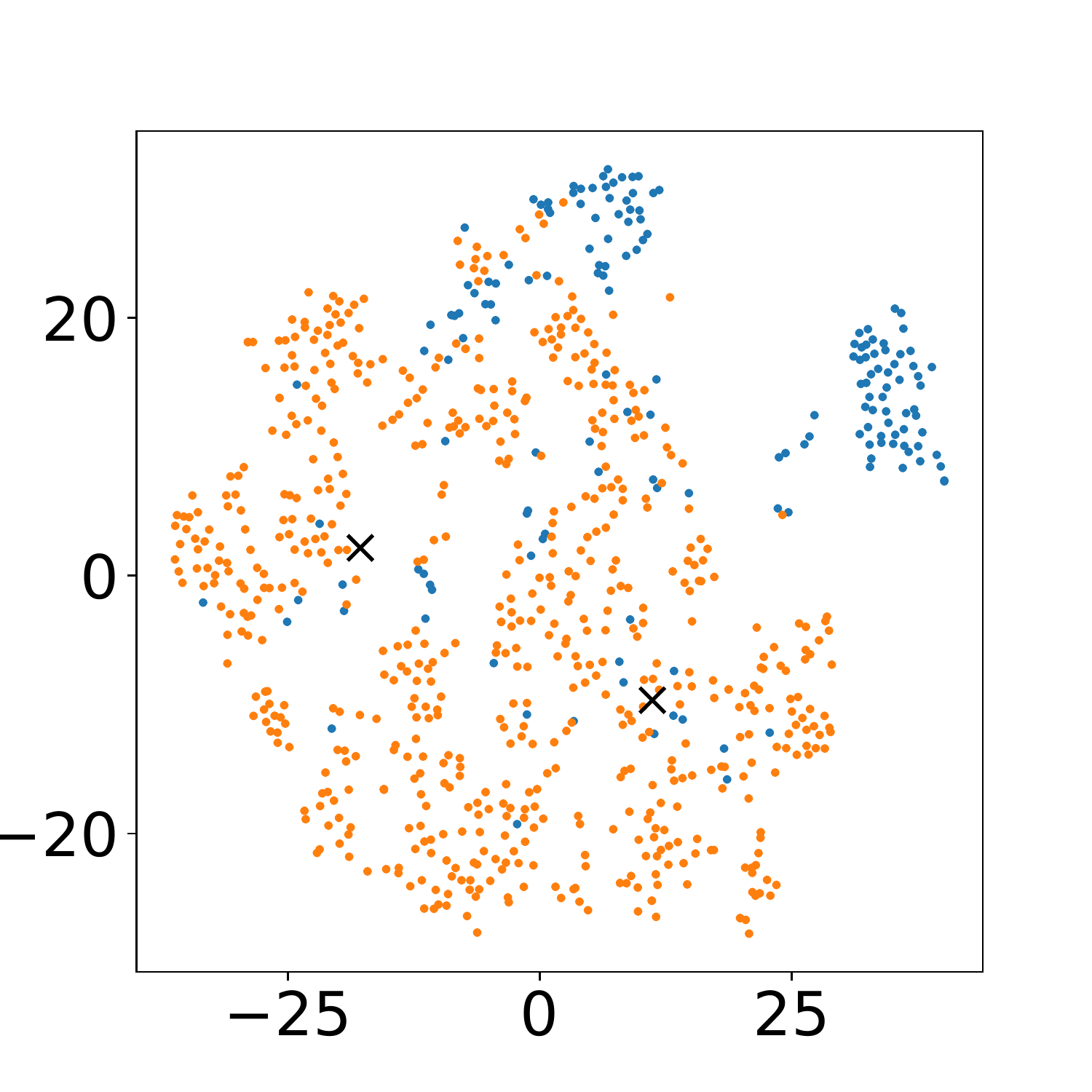}
    }
    \\
    \subfigure[PROTEIN-S]{
    \includegraphics[height=6cm]{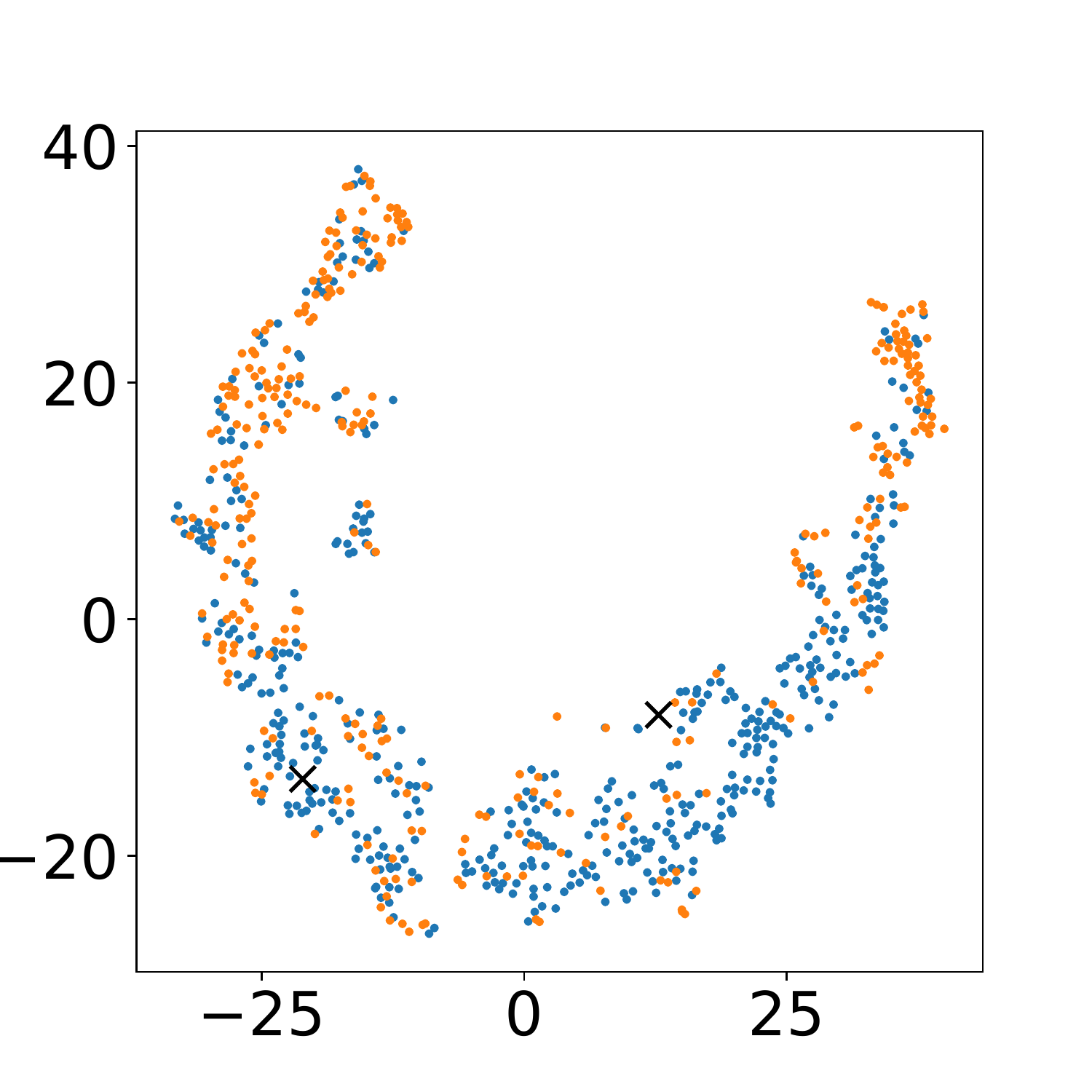}
    }
    \hspace{-3mm}
    \subfigure[PROTEIN]{
    \includegraphics[height=6cm]{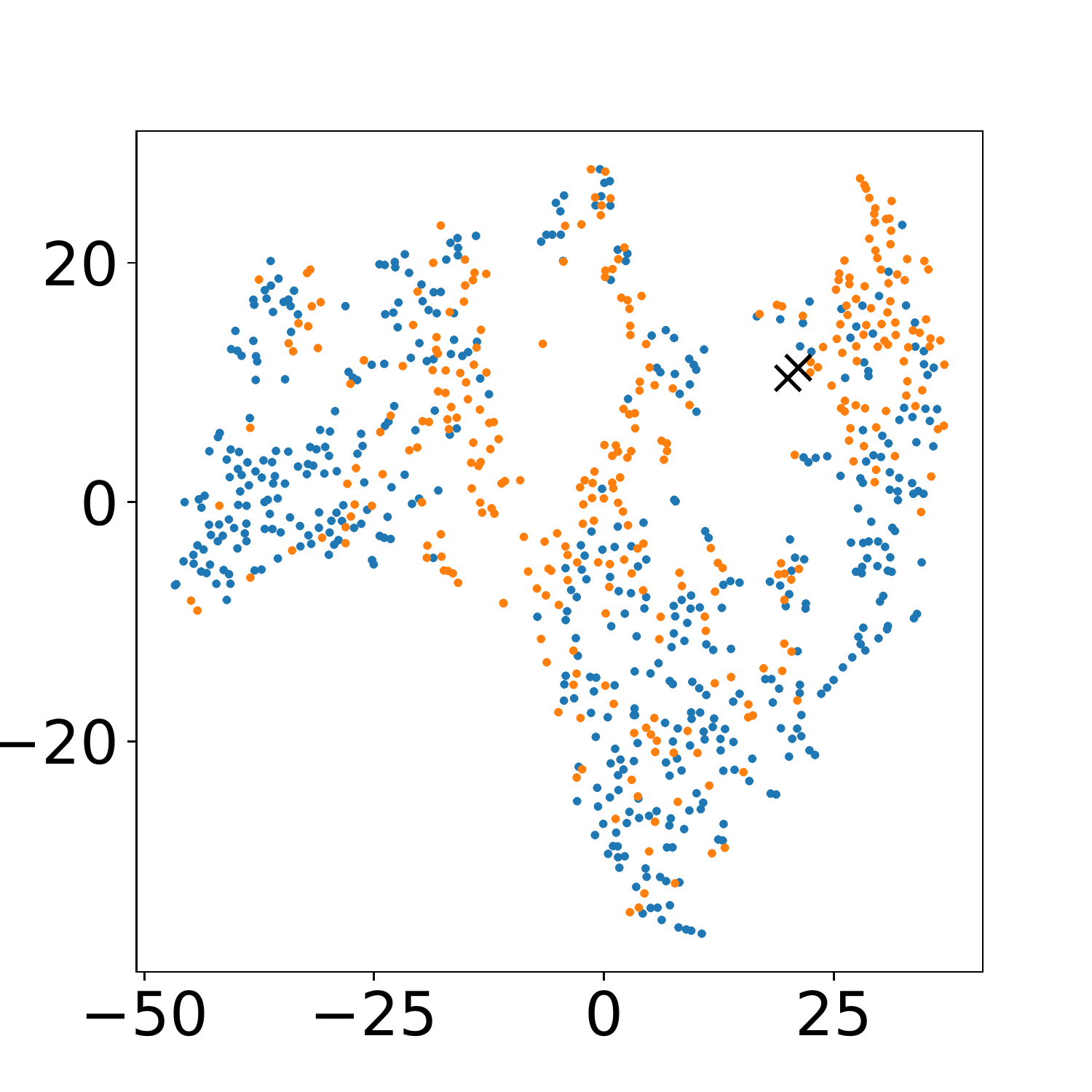}
    }
    \caption{(Good cases) The t-SNE plots of the learned latent representations for some representative datasets. The colors of the points indicate the real categories of the representations. The black crosses indicate the centers of Gaussian components of the prior distribution.}
    \label{fig:tsne1}
\end{figure}

\subsection{Initialization of graph/graphon factors}
The GWF and the GDL learn $C$ graph factors, respectively.
Their factors can be explained as representative adjacency matrices.
Our GNAEs learn $C$ graphon factors, each of which is explained as the step function induced from a representative graph. 
Both the GNAEs and the GWF allow the factors to have different sizes (numbers of partitions), so we leverage $C$ observed graphs to initialize their factors. 
The GDL requires the factors to have the same size, so we initialize its factors as random matrices, whose sizes are equal to the average size of the factors used in the GWF.

\begin{figure}[t]
    \centering
    \subfigure[PTC-MR]{
    \includegraphics[height=6cm]{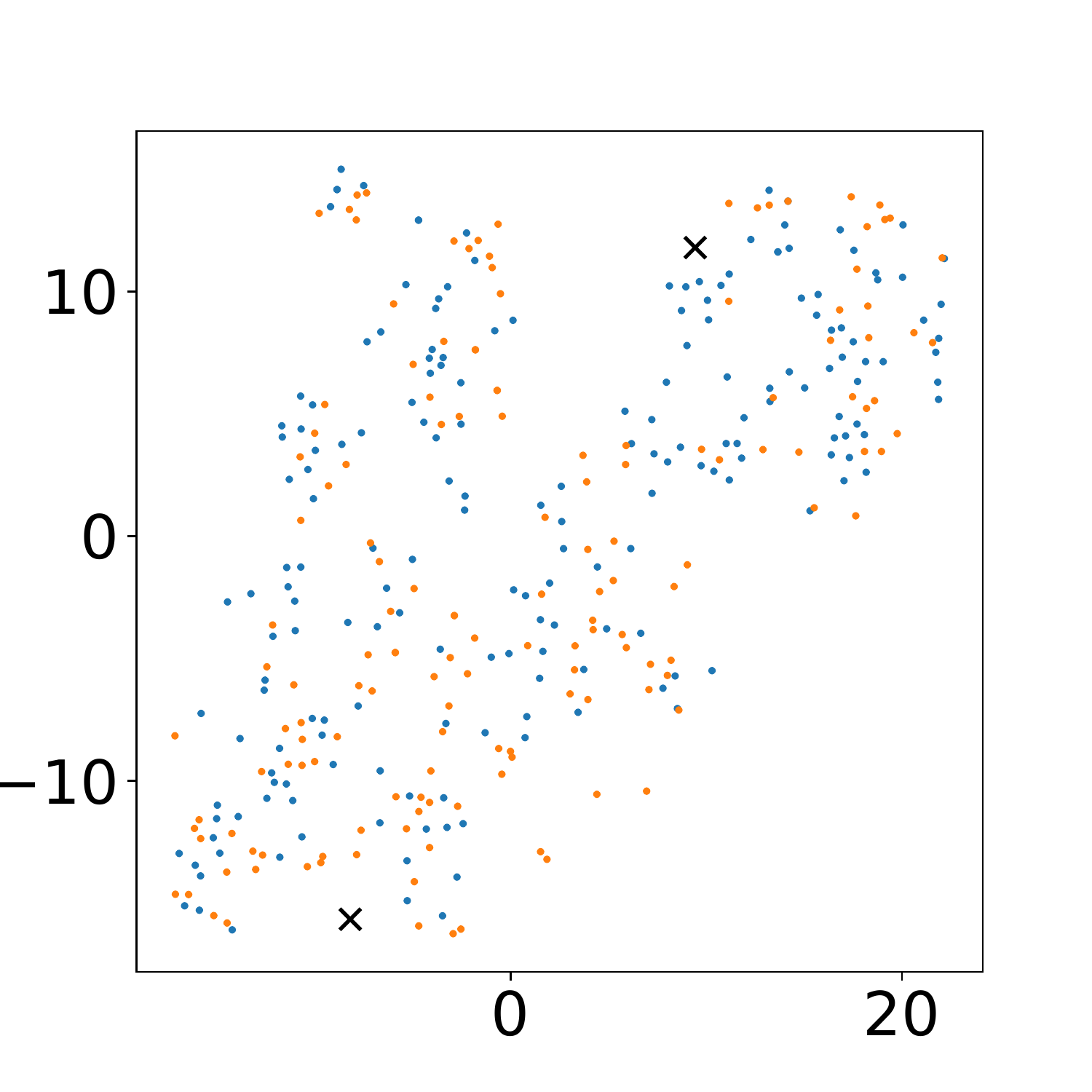}
    }
    \hspace{-3mm}
    \subfigure[ENZYMES]{
    \includegraphics[height=6cm]{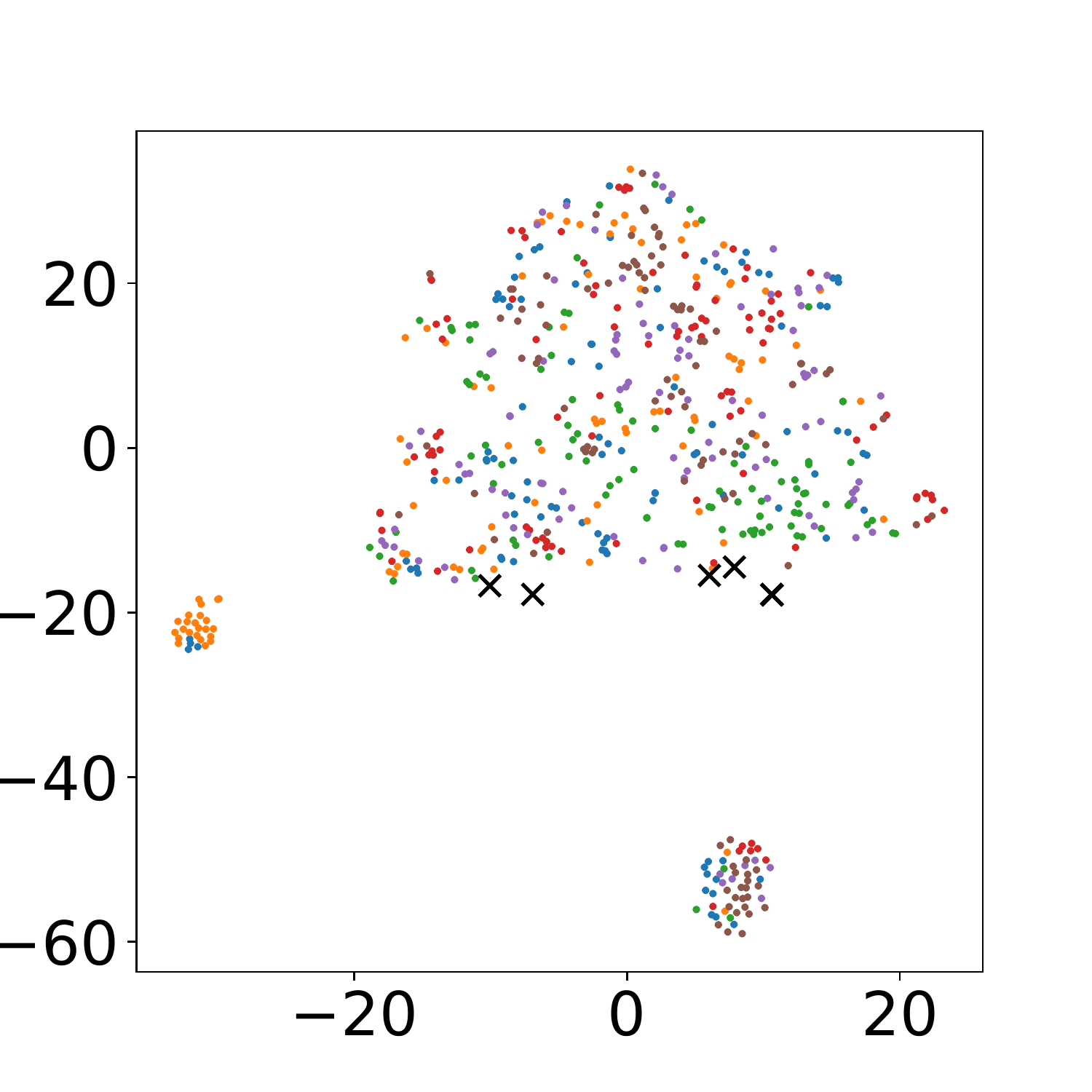}
    }\\
    \subfigure[IMDB-B]{
    \includegraphics[height=6cm]{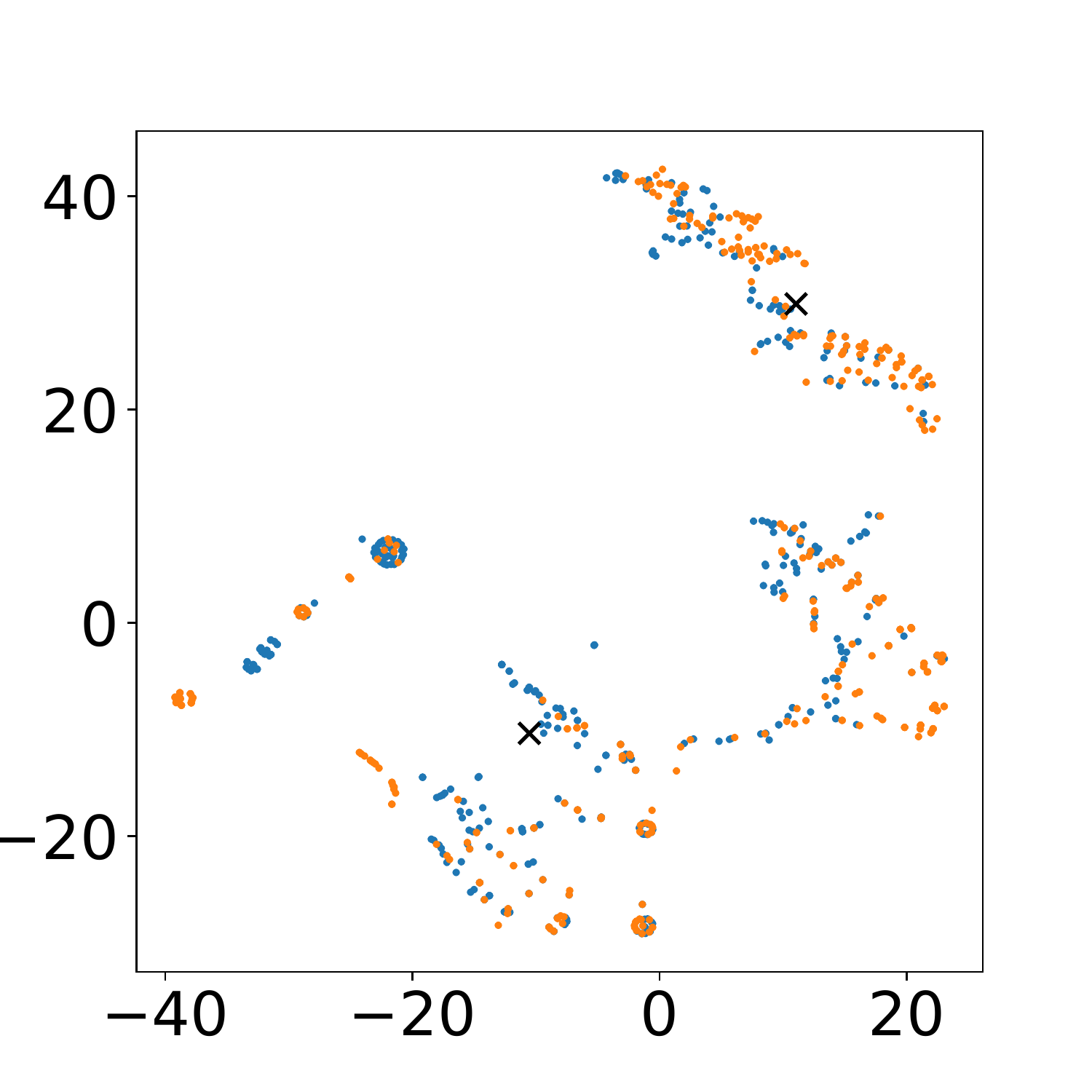}
    }
    \hspace{-3mm}
    \subfigure[IMDB-M]{
    \includegraphics[height=6cm]{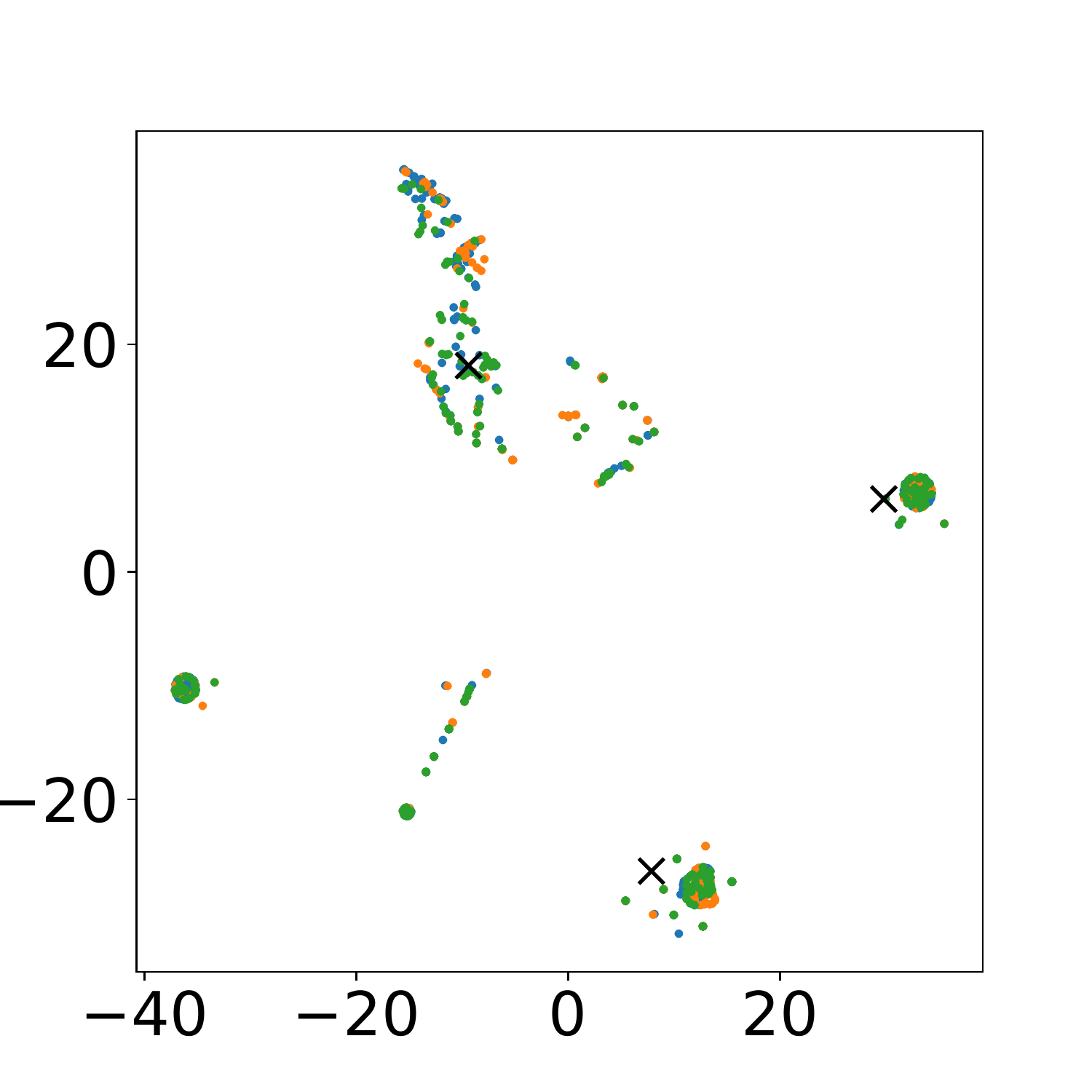}
    }
    \caption{(Bad cases) The t-SNE plots of the learned latent representations for some representative datasets. The colors of the points indicate the real categories of the representations. The black crosses indicate the centers of Gaussian components of the prior distribution.}
    \label{fig:tsne2}
\end{figure}

\subsection{Visualizations of the learned graph representations}
Besides the six datasets reported in the main paper, we consider two more datasets: the AIDS dataset~\cite{riesen2010graph}, which contains 2000 molecules that are active or inactive to HIV virus; the PROTEIN-S dataset, which is the same with the PROTEIN dataset but applies simplified node attributes. 
Figures~\ref{fig:tsne1} and~\ref{fig:tsne2} show the t-SNE plots of the graph representations learned by our GNAE$_{\text{RAML}}$. 
We can find that the representations indicate obvious clustering structures in some situations.
However, in some challenging datasets, $e.g.$, the PTC-MR, the ENZYMES and the IMDB-M, our model does not work well. 
The reasons for the unsatisfying results may include: 
\begin{itemize}
    \item \text{Data sparsity} The PTC-MR just contains 344 molecules, so our model has a high risk of overfitting.
    \item \text{Multi-class} The graphs in the ENZYMES and the IMDB-M belong to multiple classes, which increases the difficulty of clustering.
\end{itemize}

\end{document}